%% file: main.tex
\RequirePackage[svgnames,table]{xcolor}
\documentclass[11pt,letterpaper,logo]{yalearxiv}
\input{preamble}

\input{command}
\usepackage{capt-of}

\providecommand{\bs}[1]{\boldsymbol{#1}}
\title{Awomo-SimDataEngine: Agentic Simulation-Ready World Generation}
\runningtitle{Awomo-SimDataEngine: Agentic Simulation-Ready World Generation}
\author{Awomo-PhysicalRSI Team\\
  {\small Project page: \url{https://awomo-westlakedi.github.io/Awomo-SimDataEngine/}}\\}
\hypersetup{colorlinks=true,linkcolor=WestlakeBlue,citecolor=WestlakeBlue,urlcolor=WestlakeBlue}
\begin{document}
\begin{abstract}
\input{sections/00_abstract}
\end{abstract}
\maketitle

\input{sections/01_introduction}
\input{sections/02_related_work}
\input{sections/03_overview}
\input{sections/04_asset_generation}
\input{sections/05_scene_generation}
\input{sections/06_harness}
\input{sections/07_policy_generation}
\FloatBarrier
\input{sections/09_conclusion}

\section*{Contributors}
All core contributors listed below contributed \textbf{equally} to this work and are listed \textbf{alphabetically by first name}:

\textbf{Data Infra \& Model Training}: Danjiao Ma, Enhui Ma, Haohan Liu, Heng Jia, Hui Shan, Jiahuan Zhang, Jiangdi Xu, Kaiwen Guo, Linwei Zhang, Liyang Jin, Maochun Luo, Pengyao Niu, Shiwen Li, Shuangyu Feng, Tianheng Wang, Xin Wang,
  Yongqiang Huang, Zhaozhi Wang, Zijian Ma

\textbf{Project Leads}: Jianhua Xu, Tong Zhang, Xiangru Huang

\textbf{Project Sponsors}: Jianhua Xu, Kaicheng Yu

\section*{Acknowledgments}
We thank the contributors who supplied the image-conditioned scene and robot-policy modules. Their detailed algorithmic descriptions and experimental results are maintained by the respective module owners.

\bibliographystyle{plainnat}
\bibliography{main}

\appendix
\clearpage
\input{sections/appendix_harness}
\clearpage
\input{sections/appendix_isart}
\clearpage
\input{sections/appendix_unravel}
\clearpage
\input{sections/appendix_policyforge}
\clearpage
\input{sections/appendix_results}
\end{document}

%% file: preamble.tex
\usepackage{graphicx}
\usepackage{microtype}
\usepackage{natbib}
\setcitestyle{square}
\usepackage{subcaption}
\usepackage{booktabs}
\usepackage{amsmath}
\usepackage{amssymb}
\usepackage{mathtools}
\usepackage{amsthm}
\usepackage{enumitem}
\usepackage{multirow}
\usepackage{tabularx}
\usepackage{array}
\usepackage{longtable}
\usepackage{xcolor}
\usepackage{url}
\usepackage{listings}
\usepackage{tcolorbox}
\usepackage{tikz}
\usetikzlibrary{arrows.meta,positioning,fit,calc,backgrounds}
\usepackage{adjustbox}
\usepackage{xspace}
\usepackage{ragged2e}
\usepackage{pifont}

\lstdefinestyle{promptstyle}{
  basicstyle=\ttfamily\footnotesize,
  breaklines=true,
  breakautoindent=false,
  breakindent=0pt,
  postbreak=\mbox{\textcolor{gray}{$\hookrightarrow$}\space},
  columns=fullflexible,
  keepspaces=true,
  frame=single,
  framesep=5pt,
  xleftmargin=6pt,
  xrightmargin=6pt,
  aboveskip=8pt,
  belowskip=8pt,
  showstringspaces=false,
}

%% file: sections/00_abstract.tex
\begin{center}
    \includegraphics[width=0.98\linewidth]{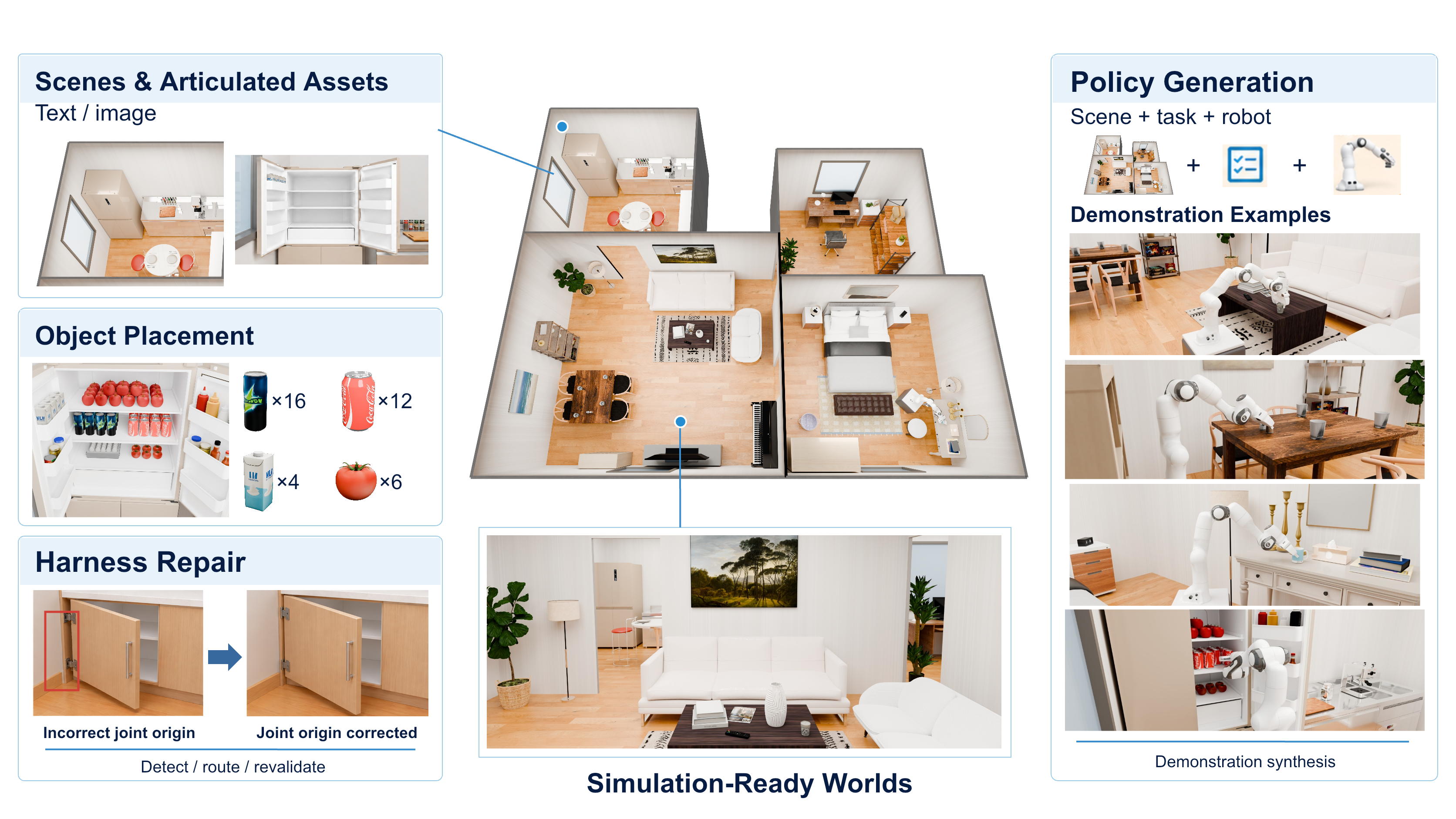}
\end{center}
\vspace{0.6em}
Generating useful robot-training data requires more than visually plausible
scenes: objects must support interaction, placements must remain physically
valid, and tasks must admit repeatable execution. We present
\textbf{Awomo-SimDataEngine}, an agentic system that connects asset and scene
generation to robot demonstration synthesis. Shared asset services provide
rigid and articulated objects, including structure-grounded part and joint
generation with ISArt. Scene generation supports two complementary routes:
Unravel reconstructs editable scenes from images, while SimForge builds
single-room and multi-room environments from text. A graph-native harness
coordinates construction, validation, and
bounded repair, routing failures to the responsible module while retaining
unaffected scene state. PolicyForge binds validated worlds to tasks and robot
embodiments to produce replayable demonstrations. Evaluations cover asset
geometry, scene quality, and downstream policy learning. On MuJoCo-based
LIBERO-Plus, co-training with Isaac Sim demonstrations improves the overall
success rate of a World-Action Model (WAM) from $77.17\%$ to $89.43\%$. Goal and spatial
success improve by $31.66$ and $6.25$ percentage points, respectively.
These results support the utility of the generated data for cross-simulator
policy training, with more limited gains on long-horizon tasks.

%% file: sections/01_introduction.tex
\section{Introduction}
\label{sec:intro}

Robot learning depends on demonstrations that cover varied objects,
environments, and manipulation goals. Collecting such coverage in the real
world is costly, while manually preparing each simulated environment limits
the scale and diversity of synthetic data. Generative models offer a way to
expand this training distribution, but their outputs must support interaction:
a plausible image or textured mesh can still hide invalid collision geometry,
inaccessible interiors, incorrect joints, unstable placements, or an
unreplayable initial state. These defects affect both whether demonstrations
can be collected and what policies learn from them.

Recent progress spans image-conditioned 3D assets, agentic scene construction,
and automated demonstration synthesis
\citep{hunyuan3d22025tencent,sam3dteam2025sam3d3dfyimages,scenesmith,sage,mimicgen,simfoundry}.
Connecting these capabilities requires agreement on object identities,
coordinate frames, physical representations, task requirements, and validation
results. An asset replacement changes the geometry available for registration;
a revised scene changes the reset and interaction conditions seen by a
controller. An integrated system must preserve these dependencies and return
failures to the stage that can resolve them.

\begin{figure}[t]
    \centering
    \includegraphics[width=\linewidth]{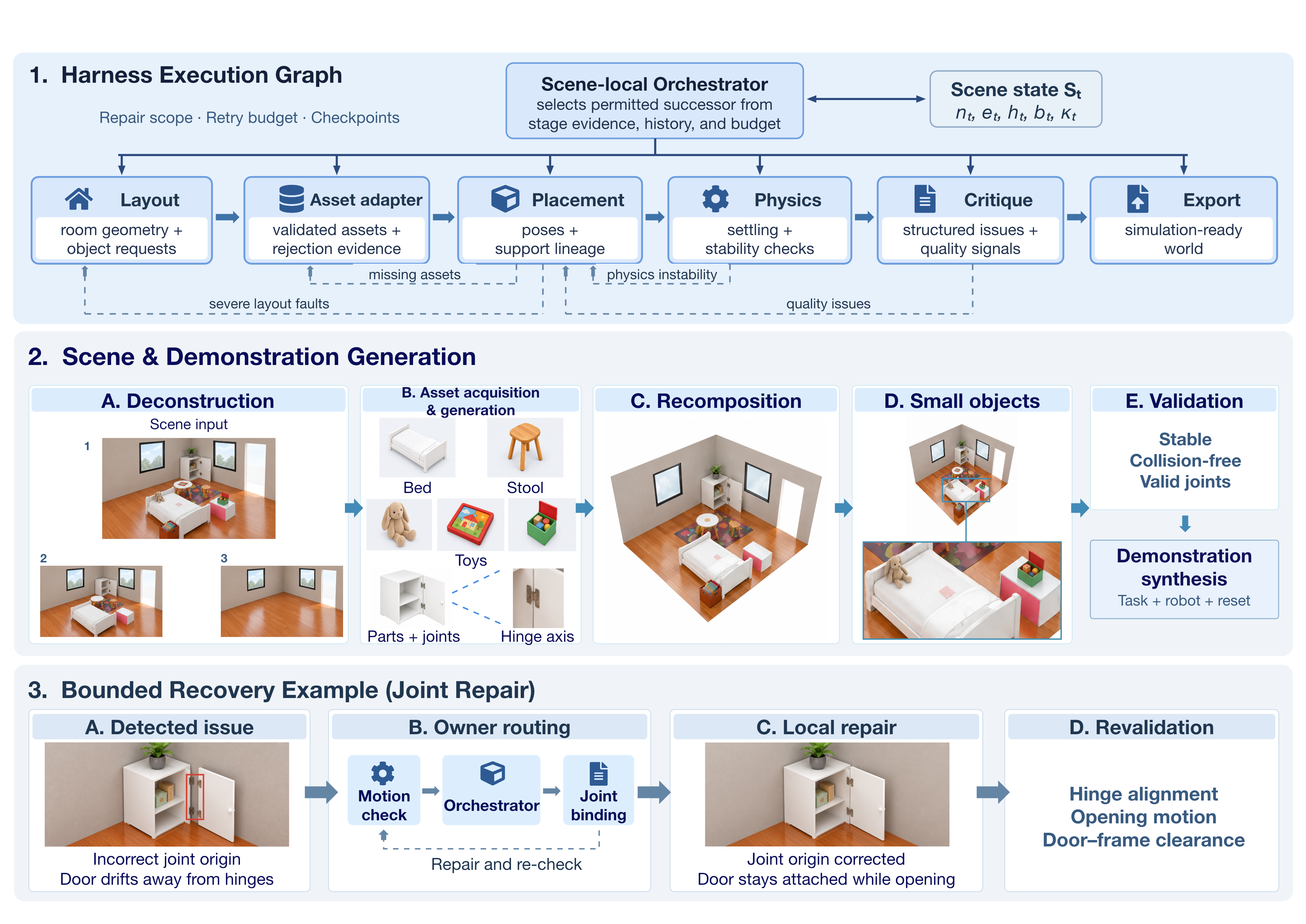}
    \caption{\textbf{Awomo-SimDataEngine overview.}
    (1) The graph-native harness coordinates execution, validation, and bounded repair.
    (2) The illustrated image-driven route uses Unravel to reconstruct scenes
    using acquired or generated assets, including
    articulated assets generated by ISArt.
    PolicyForge synthesizes demonstrations from validated worlds with task, robot,
    and reset specifications. (3) A schematic example illustrates failure detection
    and owner routing, followed by joint-origin correction and motion revalidation.}
    \label{fig:simforge-overview}
\end{figure}

We introduce \textbf{Awomo-SimDataEngine}, an agentic system for constructing
simulation-ready worlds and converting them into robot demonstrations.
The system connects asset generation, scene generation, and demonstration
synthesis through a \emph{Graph-Native Harness} for execution and repair.
Scene generation supports both visual references and language descriptions:
\emph{Unravel}~\citep{shan2026unravelrecoveringinteractive3d} reconstructs editable scenes from single-view images, whereas
\emph{SimForge} builds single-room and multi-room environments from text.
Both routes use shared asset services and produce worlds that can be checked,
edited, and bound to robot tasks. Figure~\ref{fig:simforge-overview}
illustrates the shared execution framework and the image-driven route.

\paragraph{From assets to interactive scenes.}
Scene construction begins with assets that are usable beyond rendering.
The asset module combines retrieval and generation with common geometry,
collision, and physical checks (Section~\ref{sec:asset}). Articulated objects
add a structural requirement: their parts and joints must describe an
executable mechanism. \emph{ISArt} addresses this through geometric--kinematic
planning, structure-conditioned part generation, and articulation binding.

The two scene-generation routes use these assets differently
(Section~\ref{sec:scene}). Unravel progressively removes objects from a
reference image to expose occluded regions, then uses accumulated point-map
evidence to register and reassemble independently editable assets. SimForge
instead plans room geometry, object requirements, and spatial arrangements
from language, supporting both individual rooms and multi-room houses.
For both routes, plausible appearance
alone does not establish valid support, collision geometry, or articulation.

\paragraph{From validated worlds to demonstrations.}
Construction failures need not be repaired where they are detected. A missing
asset calls for acquisition; an intersecting object calls for a placement
change; an incorrect joint may require revising the articulated asset.
The Graph-Native Harness (Section~\ref{sec:harness}) makes these handoffs
explicit through shared state, validation evidence, bounded retries, and
checkpoints. It retains unaffected work and revalidates the dependencies of
each edit before construction proceeds.

\textbf{PolicyForge} then binds an accepted world to a task, robot embodiment,
and reset specification (Section~\ref{sec:policyforge}). Scripted or
motion-planned teachers execute in Isaac Sim, producing synchronized images,
proprioception, and actions; successful validated episodes are retained for
training. Scene variants, object substitutions, and initial-state and language
augmentation diversify the resulting data. World validity, demonstration
success, and downstream policy performance remain distinct outcomes.

\paragraph{Experimental evidence.}
Module evaluations support the geometry and reconstruction methods: ISArt
achieves part-geometry CD $0.163$ versus $0.268$ for the combined
TRELLIS.2 + P3-SAM + X-Part baseline on 60 objects, while Unravel improves
blind human asset-fidelity scores from $4.9$ to $8.6$ on 160 objects.
These protocols are detailed in Sections~\ref{sec:isart} and~\ref{sec:unravel-evidence}.
Text-driven scene quality is evaluated separately on 179 room prompts and
31 multi-room prompts using the SceneEval and SceneSmith protocols
(Section~\ref{sec:simforge-text}). These evaluations test complementary
properties rather than a single pooled quality score.
On the MuJoCo-based LIBERO-Plus benchmark, SimData-Ultra co-training increases
World-Action Model success from $77.17\%$ to $89.43\%$ when comparing each configuration's best
evaluated checkpoint up to and including 13K steps, selected by overall
success, with gains of $31.66$ and $6.25$ percentage points in goal and
spatial generalization (Section~\ref{sec:experiments}). These results support the training utility of the
combined data-generation system across simulator boundaries; they do not
isolate each generator or augmentation's contribution. The smaller
long-horizon gain motivates richer sequential task supervision.

This report makes four contributions:
\begin{enumerate}
\item \textbf{Simulation-ready asset generation}, including ISArt's
structure-grounded articulated-object pipeline and shared candidate
validation, canonicalization, collision construction, and physical augmentation.
\item \textbf{Complementary scene-generation routes}: image-driven
reconstruction with Unravel and text-driven single-room and multi-room
construction with SimForge.
\item \textbf{Integrated execution and demonstration synthesis}: a
Graph-Native Harness coordinates attributed repair and acceptance, while
PolicyForge generates replayable demonstrations across scenes, tasks, and
robot embodiments.
\item \textbf{Cross-simulator policy-learning evidence}, showing a
$12.26$-point overall gain on LIBERO-Plus at the best evaluated checkpoints
within a 13K-step budget,
with the largest improvement in goal generalization.
\end{enumerate}

%% file: sections/02_related_work.tex
\section{Related Work}
\label{sec:related}

Awomo-SimDataEngine connects interactive asset generation, text- and image-driven scene
construction, agentic execution, and demonstration synthesis. We position
these components against the corresponding literature and distinguish
world-construction capabilities from downstream policy measurements.

\subsection{3D assets and interactive environments}
\label{sec:related-simulation}
Hunyuan3D, TRELLIS, and SAM 3D generate geometry and appearance from visual
conditions
\citep{hunyuan3d22025tencent,lai2025hunyuan3d25highfidelity3d,xiang2024structured,sam3dteam2025sam3d3dfyimages}.
Their outputs supply candidate assets for Awomo-SimDataEngine, which additionally checks
canonical orientation, measured extents, collision compatibility, and physical
attributes before simulator use. For articulated objects,
Section~\ref{sec:asset} introduces ISArt: structural planning conditions part
generation, and agentic binding and motion verification connect the parts to
an executable kinematic representation.

PartCrafter and OmniPart address part-level geometry generation, while
Articulate AnyMesh, Articulate-Anything, and SPARK estimate or reconstruct
articulated representations
\citep{isart_partcrafter,isart_omnipart,isart_articulateanymesh,isart_articulatea,isart_spark}.
ISArt connects explicit structural planning to a learned multi-part generator
and then refines articulation through simulation feedback. Its fixed-geometry
binding comparison separates joint estimation from the quality of generated
meshes; the part-generation comparison evaluates the latter independently.

HSSD-200, RoboCasa, and BEHAVIOR-1K provide reusable assets, scenes, or
household interaction settings \citep{hssd200,robocasa,behavior1k}. They
motivate the interaction requirements that generated assets must satisfy.
Awomo-SimDataEngine exposes retrieval and generation through a common asset lifecycle,
preserving source provenance separately from instance poses, articulation
states, and simulator-specific packaging.

\subsection{Scene synthesis and single-view construction}
\label{sec:related-layout}
ATISS and DiffuScene learn distributions over indoor object arrangements;
InstructScene, LayoutGPT, and PhyScene incorporate instructions, structured
relations, or physical constraints
\citep{atiss,diffuscene,instructscene,layoutgpt,physcene}. These approaches
provide useful priors for scene composition. Awomo-SimDataEngine's image-driven
path follows Unravel~\citep{shan2026unravelrecoveringinteractive3d} (Section~\ref{sec:scene}), using object removal to expose
occluded regions and fixed-camera point evidence to guide reconstruction and
recomposition. Its execution graph records stage dependencies; removal order
provides local occlusion evidence rather than a global support or depth order.

\label{sec:related-multimodal}
Text2Room, GraphDreamer, DreamScene, AnyHome, WonderWorld, and MIDI explore text-,
graph-, or image-conditioned 3D scene construction
\citep{text2room,graphdreamer,dreamscene,anyhome,wonderworld,huang2024midi}. Awomo-SimDataEngine uses
SimForge for text-driven single-room and multi-room construction, and Unravel
for single-image reconstruction. The latter retains object identities, masks,
point provenance, and registration evidence. Settling and visual audit check
the recomposed scene before packaging. This image-driven route does not imply
multi-view reconstruction or uniquely recoverable
metric geometry from an unconstrained image.

\subsection{Agentic generation and execution harnesses}
\label{sec:related-agents}
ReAct, Reflexion, Tree of Thoughts, and Graph of Thoughts make tool use,
feedback, or structured reasoning explicit
\citep{react,reflexion,treeofthoughts,graphofthoughts}. Holodeck, SceneCraft,
SceneWeaver, SceneSmith, SAGE, and SceneCode connect language-model reasoning
to 3D scene tools or executable representations
\citep{holodeck,scenecraft,sceneweaver,scenesmith,sage,scenecode}.

Awomo-SimDataEngine coordinates specialized modules through the execution contract in
Section~\ref{sec:harness}. Each repair records typed state, evidence, the
responsible module, affected workset, remaining budget, and checkpoint
decision. The contract spans ISArt articulation refinement, Unravel editing
and reconstruction, and world-owned PolicyForge feedback. Preserved state is
reused only when its input dependencies remain valid. These mechanisms support
recovery within a run; the policy experiments do not independently establish
harness efficiency or learning in the orchestrator.

\subsection{Integrated worlds and policy-data synthesis}
\label{sec:related-policy}
EmbodiedGen V2 connects simulation-ready assets, affordances, task-driven
worlds, editing, and simulator deployment \citep{embodiedgenv2}. SimFoundry
reconstructs digital twins from video and generates object, scene, and task
cousins for policy learning and evaluation \citep{simfoundry}. WorldComposer
also connects visual reconstruction and digital cousins to robot learning
\citep{worldcomposer}. These systems demonstrate the importance of linking
world representations to interaction and downstream training.

\begin{table}[t]
\centering
\small
\caption{System positioning. Conditions, construction procedures, and
evaluation settings differ; this comparison describes scope rather than a
matched performance ranking.}
\label{tab:multimodal-comparison}
\begin{tabularx}{\linewidth}{@{}p{0.17\linewidth}XX@{}}
\toprule
System & World construction & Connection to policy learning \\
\midrule
EmbodiedGen V2 & Simulation-ready assets, affordances, task-driven assembly,
and editable worlds & Simulator deployment and embodied policy training \\
SimFoundry & Video-based digital twins and object, scene, and task cousins
& Real-to-sim policy evaluation and sim-to-real learning \\
Awomo-SimDataEngine & ISArt-assisted assets, Unravel image reconstruction, and
SimForge text-driven scenes, coordinated by a shared harness & PolicyForge
demonstrations in Isaac Sim; measured co-training utility on MuJoCo benchmarks \\
\bottomrule
\end{tabularx}
\end{table}

MimicGen expands demonstrations from a small set of human trajectories,
RoboMimic studies learning from offline demonstrations, and Diffusion Policy
models visuomotor actions through diffusion
\citep{mimicgen,robomimic,diffusionpolicy}. PolicyForge complements this work
by preparing worlds and task contracts for scripted or motion-planned
teachers, recording replayable resets and synchronized observations and
actions, and retaining successful demonstrations. Its contribution is the
generation and validation pipeline rather than a new policy optimizer.

SceneEval and RelScene evaluate semantic coherence and spatial relations
\citep{sceneeval,relscene}. World-level checks and policy success answer
different questions. Section~\ref{sec:experiments} evaluates whether Awomo-SimDataEngine
data improves a WAM policy across simulator boundaries; it does not rank
generators under a common scene benchmark or establish real-world transfer.
The system connects asset structure, image-grounded scene evidence, bounded
repair, and validated demonstration synthesis in one workflow.

%% file: sections/03_overview.tex
\section{Overview}
\label{sec:representation}

Awomo-SimDataEngine connects simulation-ready asset generation, text- and image-driven scene
generation, and automated demonstration synthesis through a Graph-Native
Harness. Sections~\ref{sec:asset}--\ref{sec:scene} describe ISArt, Unravel, and SimForge;
Sections~\ref{sec:harness}--\ref{sec:policyforge} describe their orchestration
and the PolicyForge interface. Module-level asset and scene evaluations are
reported separately from downstream policy co-training results.

\paragraph{Input and scene-generation routes.}
The input is $C=\{T_{\textup{text}},I,Q\}$: a scene description, one image,
and an optional task request. At least one of text or image is supplied.
SimForge constructs scenes from text without requiring a reference image.
Unravel reconstructs scenes from one single-view image, optionally accompanied
by a description; this description may also be derived from the image.
Object-centric and optional open-state references used inside ISArt are
derived guidance for asset generation, not extra input scene views.
Without $Q$, the output is a scene-only world; otherwise a task contract can
be bound for demonstration synthesis.

\paragraph{Asset generation.}
Object requests are served by retrieval, image-to-3D models, specialized
generation, or procedural modeling. ISArt converts a single object image
into textured part meshes, a kinematic tree, and joint parameters through
structural planning, structure-grounded generation, articulation binding,
and simulation validation. All acquisition branches share candidate
validation, canonical transforms, collision construction, physical
augmentation, and asset-source provenance.

\paragraph{Scene generation.}
Unravel's training-free plan--edit--verify loop progressively removes
objects from the input scene image. Before each removal, crops and masks
guide object extraction. Fixed-camera point maps are aligned using unedited
anchors, fused only where new surfaces are exposed, and attributed to
object instances by pixel and removal-round provenance. Recomposition
starts from the room shell and proceeds in reverse removal order, followed
by constrained registration, support-aware settling, and visual audit.
The reconstructed scene contains independently editable assets and their
instance transforms. SimForge instead plans room geometry and object placement
from text, supporting single-room and multi-room scenes. Both routes use
shared asset services and undergo validation before downstream use.

\paragraph{Graph-Native Harness.}
The execution graph tracks typed state, validation evidence, the responsible
module, repair scope, attempt budget, and checkpoints. Stage dependencies
determine which records become invalid after an edit. Hard validation and
simulator packaging govern world acceptance; a saved render or mesh is
insufficient. The harness coordinates module-specific repair while retaining
the provenance needed for replay.

\paragraph{Policy generation and evaluation.}
PolicyForge binds an accepted world to a task, robot embodiment, and reset
specification. Scripted or motion-planned teachers produce synchronized
observations, states, and actions; successful episodes form demonstrations.
Scene and task cousins, with initial-state and language augmentation,
diversify policy-training data. The measured downstream endpoint in this
report is WAM co-training in Isaac Sim-generated environments and evaluation
on MuJoCo-based LIBERO benchmarks. World construction, episode success,
and benchmark policy success remain distinct validation outcomes.

%% file: sections/04_asset_generation.tex
\section{Asset Generation}
\label{sec:asset}

Asset generation converts an object request into a validated simulator asset.
Given text, an image, or both, the module obtains candidates through retrieval,
image-to-3D generation, specialized generation, or procedural modeling.
Object-centric references remove distracting scene content and provide
appearance and structural cues. Each branch returns geometry and available
semantics, orientation, scale, articulation, and physical information to the
shared validation process in Section~\ref{sec:harness}.

\paragraph{Acquisition branches.}
CLIP-based retrieval selects candidates from reusable asset databases.
Image-to-3D models such as Hunyuan3D, TRELLIS.2, and SAM 3D provide static
geometry from object-centric references or scene images and masks
\citep{hunyuan3d22025tencent,isart_trellis2,sam3dteam2025sam3d3dfyimages}.
Specialized branches handle assets requiring additional structure.
For articulated objects, our \emph{ISArt} pipeline generates part meshes
and executable kinematics jointly with agentic structural reasoning.
Retrieved or procedurally modeled objects remain alternatives when
generation fails the required checks.

\subsection{Single-Image Articulated Assets with ISArt}
\label{sec:isart}

\begin{figure}[t]
\centering
\includegraphics[width=\linewidth]{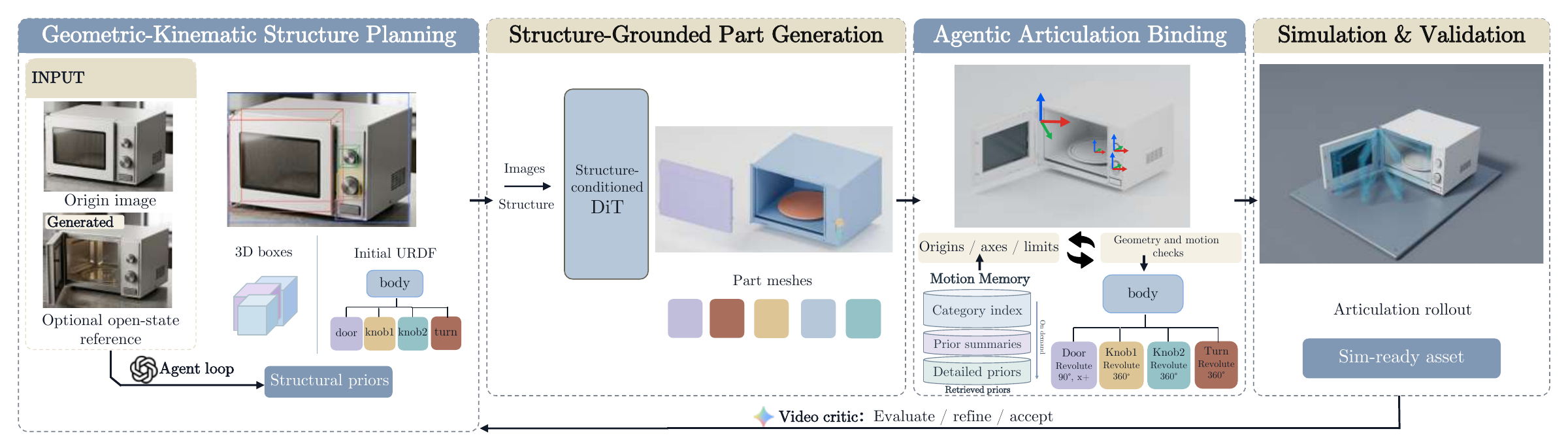}
\caption{\textbf{ISArt articulated-asset generation.} Structural planning
supplies part and kinematic priors for learned geometry generation.
Agentic binding constructs a URDF, and rendered simulation feedback guides
parameter repair, structural re-planning, or geometry regeneration.}
\label{fig:articulated-asset-generation}
\end{figure}

Given one object-centric image $I_0$, ISArt produces
\begin{equation}
\mathcal{A}=F(I_0)=(\mathcal{M},\mathcal{K},\Theta),
\qquad \mathcal{M}=\{M_i\}_{i=1}^{N},
\end{equation}
where $\mathcal{M}$ contains textured part meshes, $\mathcal{K}$ is the
URDF kinematic tree, and $\Theta$ contains joint origins, axes, and ranges.
The pipeline has four stages: geometric--kinematic structure planning,
structure-grounded part generation, agentic articulation binding, and
simulation validation (Figure~\ref{fig:articulated-asset-generation}).

\paragraph{Geometric--kinematic structure planning.}
A VLM predicts object category, part count, semantics, parent links,
joint types, coarse axes, part bounding boxes, and cavity boxes.
The joint vocabulary comprises fixed, revolute, prismatic, and continuous
joints; coarse axes use six canonical directions. A synthesized open-state
reference exposes possible concealed interiors when needed. Coarse URDF
prediction is iteratively critiqued, while part and cavity boxes are refined
through estimate--project--critique updates with the static-body box held
fixed. The resulting structure remains editable for later fallback.

\paragraph{Structure-grounded part generation.}
ISArt adapts a pretrained TripoSG geometry prior with a multi-part DiT
\citep{isart_tripo}. Each predicted part has its own latent group.
Frozen visual encoders process the original and optional open-state image;
seven part-aligned tokens encode category, semantics, joint attributes,
parent relation, box center, box size, and cavity attributes.
Two mechanisms condition denoising: \emph{E} injects a pooled part-structure
embedding into every DiT layer, and \emph{C} cross-attends from geometry
tokens to the full structural token sequence. The backbone retains
within-part and across-part attention and visual cross-attention.
A pretrained decoder produces each part mesh, followed by part-level texture
synthesis. Structural modules and attention LoRA parameters are optimized
with the rectified-flow objective, with structural residuals zero-initialized.
Appendix~\ref{app:isart} details the tokenization and conditioning equations.

\paragraph{Articulation binding and simulation validation.}
Part identifiers bind meshes to URDF links. Parent relations and coarse
joint predictions initialize the tree; geometry and retrieved category-level
motion priors guide refinement of axes, pivots, and ranges. ISArt's
hierarchical articulation memory can expand a motion summary into detailed
joint guidance. This module-specific memory is separate from the harness's
run-local checkpoint state. In Isaac Sim, joint motions are rendered and
a video critic checks direction, pivot, range, interference, and collisions.
Feedback drives parameter edits, rollback, and retry; persistent failures
escalate to structural planning or geometry regeneration. Invalid candidates
remain rejected rather than silently becoming static objects.

\paragraph{Module-level evidence.}
On 60 articulated objects from two sources spanning 12 categories, ISArt
reports part-geometry Chamfer Distance $0.163$, compared with $0.268$ for
the TRELLIS.2 + P3-SAM + X-Part pipeline~\citep{isart_trellis2,isart_p3sam,isart_xpart}, and F-Scores $0.604$ and $0.789$
at thresholds $0.05$ and $0.1$. With fixed ground-truth part geometry,
ISArt's URDF binding has AxisErr $0.07$, PivotErr $0.05$, TypeErr $13.33\%$,
and reported binding success $100\%$. This last measurement evaluates
binding with supplied geometry, not end-to-end asset generation.
Complete comparisons and training details are in
Appendices~\ref{app:isart} and~\ref{app:results-isart}.

\subsection{Standardization, Physical Augmentation, and Reuse}
\label{sec:asset-standardization}

Accepted geometry is transformed into a declared canonical frame, retaining
the source-to-canonical transform. Orientation and scale are estimated from
the request and reference evidence; placement extents are measured from the
standardized mesh. Collision construction uses convex decomposition and
interaction-clearance checks. Small objects use VHACD; larger objects may
be segmented and decomposed at part level. Mass, friction, and other available
physical attributes remain associated with the asset.

Visual meshes, collision proxies, articulated link frames, and simulator
representations retain auditable scale and transform relationships.
The Asset Registry records accepted and rejected candidates, sources,
validation outcomes, and reusable geometry separately from instance poses
(Appendix~\ref{app:asset-registry}). Scene registration and support validation
operate on the instantiated assets in Section~\ref{sec:scene}; target-specific
load and package checks complete the acceptance process.

%% file: sections/05_scene_generation.tex
\section{Scene Generation}
\label{sec:scene}

Awomo-SimDataEngine supports two parallel scene-generation routes.
\emph{Unravel} reconstructs an editable scene from a single image, optionally
accompanied by a description; \emph{SimForge} constructs single-room scenes
and multi-room houses from text. Neither route requires the other
as a preprocessing stage. Both use the asset services in
Section~\ref{sec:asset}, retain object and articulation identities, and
produce worlds for validation and downstream demonstration synthesis.

\subsection{Image-to-Scene Generation: Unravel}
\label{sec:unravel-image}

\begin{figure}[t]
\centering
\includegraphics[width=0.95\linewidth]{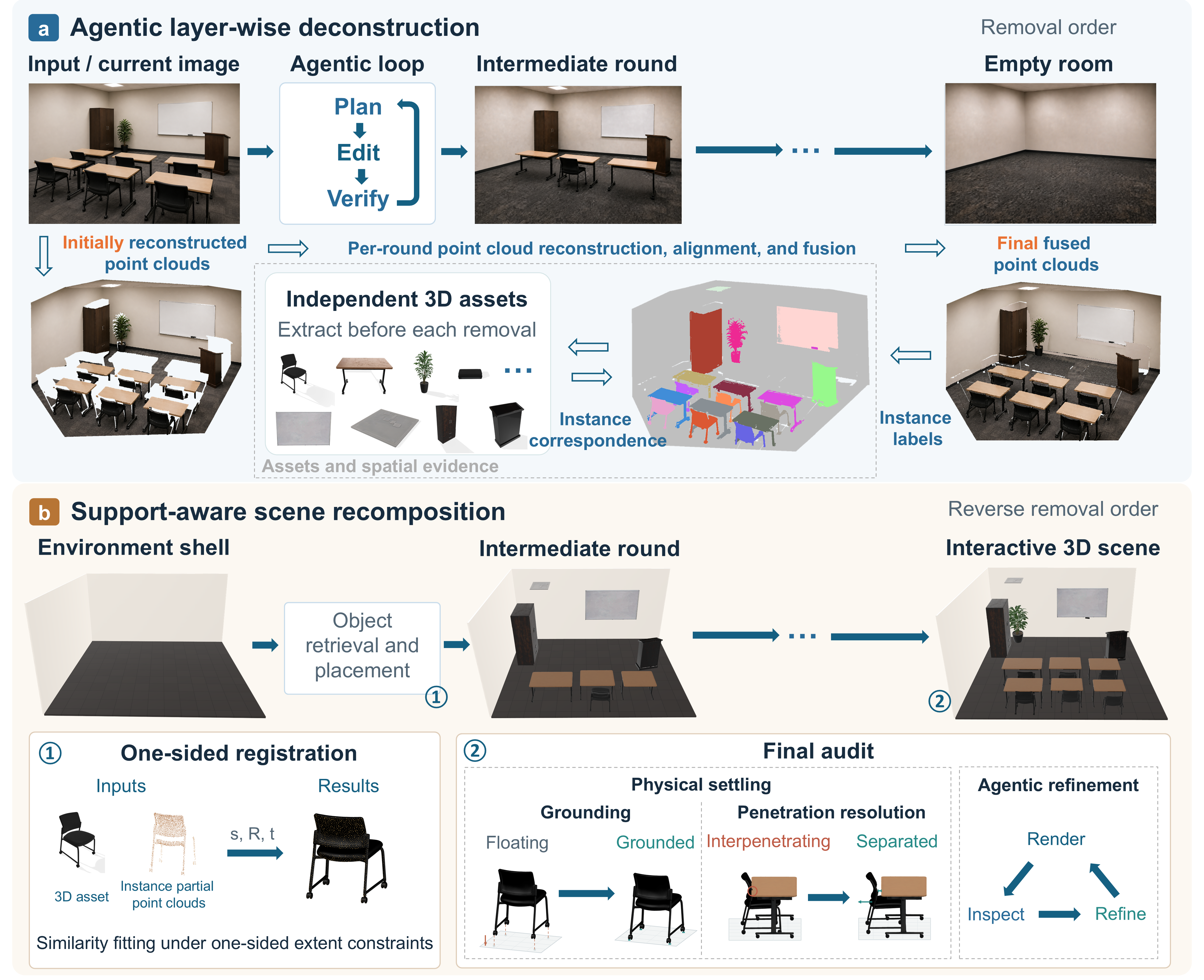}
\caption{\textbf{Unravel single-view scene generation.} Verified removals
expose objects and collect assets and point evidence. Reverse-order
recomposition registers each asset to its attributed cluster, followed by
support-aware settling and a final audit.}
\label{fig:scene-generation}
\end{figure}

Unravel~\citep{shan2026unravelrecoveringinteractive3d} is a training-free scene-reconstruction procedure that
deconstructs a single-view scene image before recomposing it as an interactive
world. Its input is one scene image, with an optional description and task
constraints as specified in Section~\ref{sec:representation}.
Unravel's output is
\begin{equation}
\mathcal{W}=\bigl(\mathcal{E},\{(a_i,T_i)\}_{i=1}^{N}\bigr),
\end{equation}
where $\mathcal{E}$ is the room shell, $a_i$ is an independent asset, and
$T_i=(\mathbf{s}_i,R_i,t_i)$ stores scale, rotation, and translation.
Figure~\ref{fig:scene-generation} shows the two-stage pipeline.

\subsubsection{Verified Layer-Wise Deconstruction}
\label{sec:unravel-deconstruction}

A VLM planner selects exposed objects for removal, protects the remaining
objects, and preserves walls and the floor. Contents are removed before
containers. An image editor removes targets and inpaints exposed regions;
a verifier checks the edit before the next round. Right before removal,
each object's best-observed crop, semantic description, bounding box,
instance mask, and removal round are recorded. A canonical object image
guides the asset-generation branch in Section~\ref{sec:asset}.
Box localization uses a global prediction with independent local cross-checks
(Appendix~\ref{app:unravel-boxes}).

The fixed camera is retained across rounds. Newly inpainted content is a
generated hypothesis about hidden surfaces, not a newly acquired physical
measurement; verification and downstream geometry checks remain necessary.
Deconstruction ends when only the room shell remains or the harness records
an unresolved failure.

\paragraph{Anchored point-map fusion.}
For round $k$, a monocular estimator returns point map $P_k$ and confidence
mask $M_k$, excluding unreliable depth-boundary pixels.
Let $E_j$ denote pixels edited in round $j$. Stable anchors are valid in
both the first and current round and have never been edited:
\begin{equation}
\mathcal{A}_k=\{p:M_k(p)\wedge M_1(p),\
p\notin\textstyle\bigcup_{j<k}E_j\}.
\end{equation}
A similarity transform aligns $P_k$ directly to $P_1$ using these anchors,
with residual-based outlier rejection. Unedited pixels keep their existing
points; fusion adds valid points only from the regions exposed by the
preceding edit. This retains earlier object geometry while accumulating
newly revealed surfaces.

\paragraph{Instance attribution.}
Every fused point retains its pixel and reconstruction round. For a point
from round $k$, candidate objects are those whose masks cover its pixel
and whose removal rounds satisfy $r_i\ge k$. The earliest surviving removal
candidate identifies the frontmost object along that ray; an empty candidate
set identifies the room shell. Object points form clusters $C_i$ for
registration. This is a per-ray rule, not a global depth ordering.

\subsubsection{Registration and Support-Aware Recomposition}
\label{sec:unravel-recomposition}

Assets are restored in descending removal round, beginning with the room
shell. Contents-before-container planning favors placing supporting geometry
before its contents. This order provides an assembly prior; support is checked
against local geometry during settling rather than assumed from occlusion
alone.

\paragraph{Registration from partial observations.}
For asset surface samples $X_i$ and cluster $C_i$, transformations are ranked
by a bidirectional median Chamfer distance:
\begin{equation}
\begin{aligned}
D_i(T)=\tfrac12\bigl[
&\operatorname{med}_{x\in T(X_i)}\min_{y\in C_i}\|x-y\|_2\\
+&\operatorname{med}_{y\in C_i}\min_{x\in T(X_i)}\|y-x\|_2
\bigr].
\end{aligned}
\end{equation}
Truncated similarity ICP refines reliable correspondences. The median and
truncation reduce the influence of neighboring-object fragments and spurious
generated geometry. Along under-observed axes $\mathcal{U}_i$, visible extent
is treated as a lower bound:
\begin{equation}
\operatorname{span}_u(T_i(X_i))\ge\operatorname{span}_u(C_i),
\qquad u\in\mathcal{U}_i.
\end{equation}
A global scale and bounded object-frame axis adjustments preserve this
constraint. Well-observed axes may contract, but partial observations cannot
justify shrinking an asset below their measured span.

\paragraph{Physical settling and audit.}
Semantic relations and local support geometry distinguish mounted objects,
which stay fixed, from movable objects, which may undergo observation-constrained
translations $\delta_i$. Settling minimizes total squared displacement subject
to support, non-interpenetration, and observation-consistency constraints.
Contact and observation-fit tolerances are specified in
Appendix~\ref{app:unravel-registration}. A render-and-inspect audit proposes
local corrections, followed by renewed settling. Failed extraction,
registration, or physical checks return to the responsible bounded stage
in Section~\ref{sec:harness}; the accepted scene retains independently
editable assets and their observation and validation records.

\subsubsection{Module-Level Reconstruction Evidence}
\label{sec:unravel-evidence}

A controlled comparison on five scenes containing 160 objects uses the same
SAM 3~\citep{carion2025sam3}, TRELLIS.2, and MoGe3~\citep{kong2026moge3} components for direct reconstruction and
Unravel. Deconstruction increases mean VLM asset-fidelity scores from
$4.7$ to $8.2$ and human scores from $4.9$ to $8.6$ on a 1--10 scale.
Mean scene point count increases from $1.53$M to $2.81$M through fusion of
four to eight rounds. Direct masks cover $77.2\%$ of the Unravel mask area
on average for objects removed after the first round.
These coverage measurements use Unravel's masks and clusters as references,
and additional points may come from generated image completion; they are not
independent geometric ground truth. Shared-asset qualitative comparisons
also separate registration quality from mesh-generation quality.
Appendix~\ref{app:results-unravel} gives the full measurements, scoring
protocol, and representative comparisons.

We further assess full-scene fidelity at the image level. For each
generated scene, we first recover the photo's camera pose. We run
truncated ICP between surface samples of the generated geometry and the
first-round MoGe3 point cloud. We then render the scene with the
estimated intrinsics and compare the render with the photo. We use
three metrics. \textbf{CLIP-I} is the cosine similarity of CLIP
ViT-L/14 image embeddings and captures semantic content. \textbf{LPIPS}
is a learned perceptual distance over VGG features. \textbf{Edge IoU}
is the intersection-over-union of Canny edge maps with a two-pixel
tolerance. It captures object layout and composition. We evaluate on
\textbf{20 scenes}. Table~\ref{tab:scene-similarity} shows that Unravel
beats GPT-6 Astra on all three metrics.

\begin{table}[t]
  \centering
  \caption{Scene-level image similarity against the reference photo.
  Mean over 20 scenes. Renders use the recovered photo viewpoint.
  \textbf{Bold} marks the better value per metric. $\Delta$ is the
  relative change of Unravel over GPT-6 Astra, computed from the displayed means.}
  \label{tab:scene-similarity}
  \small
  \begin{tabular}{l ccc}
    \toprule
    Method & CLIP-I $\uparrow$ & LPIPS $\downarrow$ & Edge IoU $\uparrow$ \\
    \midrule
    GPT-6 Astra (Ultra/ExtraHigh) & 0.803 & 0.649 & 0.107 \\
    Unravel (w/ finetuned GPT-6)   & \textbf{0.854} & \textbf{0.632} & \textbf{0.156} \\
    \midrule
    $\Delta$ (rel.)   & $+6.4\%$ & $-2.6\%$ & $+45.8\%$ \\
    \bottomrule
  \end{tabular}
\end{table}

\FloatBarrier

\subsection{Text-to-Scene Generation: SimForge}
\label{sec:simforge-text}

SimForge translates a scene description into a single-room or multi-room
world. The request specifies room types, required objects, spatial relations,
and optional interaction requirements. Layout planning establishes room
geometry, openings, and object requests; an asset adapter retrieves or
generates compatible candidates. Placement instantiates these assets on
measured supports, followed by physics checks, critique, and simulator
packaging. A reference scene image is not required. In multi-room mode,
room topology and inter-room connections provide the architectural context
for room-specific furnishing and object placement.

\subsubsection{Language-Guided Construction}
The language model specifies object requirements and spatial arrangement
intent; geometry-aware routines instantiate and validate candidate poses.
Repeated small-object groups use procedural placement, while individually
constrained items retain separate pose and support assignments.

Construction uses the registered layout, asset-adapter, placement, physics,
critique, and export stages of the Graph-Native Harness
(Section~\ref{sec:harness}). Missing assets return to acquisition; placement
or support faults return to the responsible placement scope; severe layout
faults require architectural revision. Bounded routing and checkpoints retain
unaffected work while the revised candidate is revalidated. These are shared
execution mechanisms, not an additional image-reconstruction stage.

\subsubsection{Reusable Assets and Scene Variants}
Processed assets store shared geometry, appearance, collision, and articulation
metadata. Scene instances separately retain pose, support, joint state, and
task role. This separation supports local object insertion, appearance
changes, and rigid or articulated substitution. Changed dependencies are
rebound and affected supports, neighbors, and joints are rechecked before
the edited world is accepted. Accepted worlds can then be bound to tasks
and robots by PolicyForge; scene generation does not itself imply a
successful robot demonstration.

\subsubsection{Module-Level Scene-Quality Evidence}
Room-level evaluation uses 179 prompts under SceneEval, while house-level
evaluation uses 31 multi-room prompts under SceneSmith's house-level
protocol~\citep{sceneeval,scenesmith}. The two scales are reported separately.
Table~\ref{tab:simforge-scene-quality} summarizes semantic and physical
quality. Relative to SceneSmith, SimForge
improves count agreement by 6.5 percentage points at room level and 6.4 at
house level, and accessibility by 14.7 and 13.3 points, respectively.
Detected collision is 0\% and stability is 100\% under the stated checks.
These are scene-quality measurements, not generation acceptance rates or
isolated causal estimates of Harness effectiveness.

\begin{table}[t]
\centering
\caption{Text-to-scene quality under SceneEval and SceneSmith protocols~\citep{sceneeval,scenesmith}. Room-level and house-level
cohorts are kept separate. Values are reported means $\pm$ 95\% confidence
intervals where available. SAGE$^\dagger$ uses a separate 30-scene cohort
and native collision/stability definitions; it is a reference, not a matched
ranking against the other rows~\citep{sage}. \#Obj: object count;
CNT: count agreement; ATR: attribute correctness; OOR: object--object
relations; OAR: object--architecture relations; ACC: accessibility;
NAV: navigability; COL: collision; STB: stability; OOB: out-of-bounds.
SimForge is the text-driven scene module of Awomo-SimDataEngine.}
\label{tab:simforge-scene-quality}
\begingroup
\setlength{\tabcolsep}{3pt}
\small
\resizebox{\linewidth}{!}{%
\begin{tabular}{lrrrrrrrrrr}
\toprule
Method & \#Obj & CNT $\uparrow$ & ATR $\uparrow$ & OOR $\uparrow$ & OAR $\uparrow$ & ACC $\uparrow$ & NAV $\uparrow$ & COL $\downarrow$ & STB $\uparrow$ & OOB $\downarrow$\\
\midrule
\multicolumn{11}{l}{\emph{Room-level generation (179 prompts)}}\\
HSM~\citep{hsm} & $22.7\pm2.6$ & $60.6\pm4.2$ & $61.5\pm7.3$ & $30.6\pm5.7$ & $66.2\pm6.6$ & $88.9\pm1.7$ & $99.4\pm0.4$ & $20.6\pm3.0$ & $45.2\pm3.9$ & $5.5\pm1.1$\\
Holodeck~\citep{holodeck} & $23.0\pm1.4$ & $44.2\pm4.3$ & $44.4\pm7.2$ & $16.8\pm4.6$ & $38.0\pm7.1$ & $84.9\pm1.4$ & $99.6\pm0.2$ & $12.3\pm2.3$ & $31.9\pm2.7$ & $0.8\pm0.4$\\
SceneWeaver~\citep{sceneweaver} & $13.5\pm1.0$ & $41.8\pm3.9$ & $29.1\pm6.7$ & $15.3\pm4.2$ & $31.6\pm6.5$ & $77.5\pm2.7$ & $98.1\pm1.2$ & $12.5\pm2.5$ & $37.3\pm4.4$ & $0.0\pm0.0$\\
I-Design~\citep{idesign} & $13.0\pm1.0$ & $69.3\pm4.0$ & $50.3\pm7.2$ & $28.6\pm5.5$ & $66.6\pm6.7$ & $70.1\pm2.8$ & $95.9\pm1.4$ & $3.0\pm1.5$ & $60.8\pm4.6$ & $4.3\pm2.0$\\
LayoutVLM (Cur.)~\citep{layoutvlm} & $11.2\pm1.6$ & $41.0\pm4.0$ & $25.6\pm6.6$ & $14.1\pm4.0$ & $22.4\pm6.1$ & $93.8\pm1.9$ & $99.7\pm0.2$ & $25.9\pm4.4$ & $19.4\pm3.6$ & $6.2\pm2.1$\\
LayoutVLM (Obj.)~\citep{layoutvlm} & $14.2\pm1.7$ & $55.5\pm4.7$ & $34.3\pm7.0$ & $20.7\pm5.0$ & $17.4\pm5.6$ & $91.5\pm1.6$ & $98.7\pm0.8$ & $28.9\pm3.5$ & $8.1\pm1.7$ & $5.1\pm1.4$\\
SceneSmith~\citep{scenesmith} & $71.1\pm13.0$ & $82.9\pm3.4$ & $74.4\pm6.2$ & $67.6\pm5.8$ & $80.6\pm5.5$ & $83.4\pm1.5$ & $97.6\pm1.1$ & $1.2\pm0.6$ & $95.6\pm1.7$ & $0.2\pm0.4$\\
SimForge & $67.6\pm5.1$ & $89.4\pm4.2$ & $76.8\pm7.1$ & $73.1\pm5.4$ & $79.3\pm5.1$ & $98.1\pm1.4$ & $98.4\pm1.2$ & $0.0\pm0.0$ & $100.0\pm0.0$ & $0.0\pm0.0$\\
SAGE$^\dagger$~\citep{sage} & 48.2 & --- & --- & --- & --- & --- & --- & $1.9^\dagger$ & $99.9^\dagger$ & ---\\
\midrule
\multicolumn{11}{l}{\emph{House-level generation (31 prompts)}}\\
Holodeck~\citep{holodeck} & $81.3\pm18.3$ & $64.6\pm6.8$ & $43.2\pm27.6$ & $30.1\pm10.1$ & $58.4\pm6.3$ & $79.4\pm3.0$ & $99.6\pm0.2$ & $3.8\pm3.1$ & $17.9\pm5.6$ & $0.3\pm0.2$\\
SceneSmith~\citep{scenesmith} & $214.1\pm60.9$ & $83.3\pm6.8$ & $66.8\pm24.6$ & $76.6\pm9.2$ & $82.6\pm5.8$ & $79.9\pm2.0$ & $96.5\pm2.1$ & $0.9\pm0.5$ & $79.8\pm9.1$ & $0.6\pm0.3$\\
SimForge & $242.3\pm32.9$ & $89.7\pm6.4$ & $70.3\pm26.7$ & $78.9\pm10.3$ & $81.9\pm6.1$ & $93.2\pm2.7$ & $98.1\pm1.3$ & $0.0\pm0.0$ & $100.0\pm0.0$ & $0.0\pm0.0$\\
\bottomrule
\end{tabular}}
\endgroup
\end{table}

%% file: sections/06_harness.tex
\section{Graph-Native Harness for Execution and Repair}
\label{sec:harness}

The Graph-Native Harness coordinates asset generation, Unravel reconstruction,
SimForge scene construction, simulator packaging, and PolicyForge demonstration synthesis.
Each execution edge specifies its inputs, responsible module, validation
requirements, and records needed to resume or repair the result.
Figure~\ref{fig:simforge-overview} shows these system connections. The
execution graph organizes module calls and recovery for both SimForge's
text-driven construction and Unravel's image-grounded recomposition.

\paragraph{Shared state and inputs.}
The input is $C=\{T_{\textup{text}},I,Q\}$ as defined in
Section~\ref{sec:representation}. SimForge receives a scene description;
Unravel receives one scene image with an optional description.
When $Q$ is present, its entities, interaction
requirements, reset assumptions, and success predicate are bound to the
PolicyForge task contract $\mathcal{T}$
(Section~\ref{sec:policyforge-task-contract}). Scene-only requests carry no
policy task. State records include asset candidates and canonical transforms,
ISArt part and joint identities, SimForge layouts and placement records,
Unravel masks and removal rounds,
point-cloud provenance, registrations, physical checks, export manifests,
retry budgets, and checkpoints. World versions link these records to
downstream reset and rollout records.

\paragraph{Validation at module boundaries.}
Asset validation checks semantic match, usable geometry, orientation and
scale, collision clearance, and required articulation
(Section~\ref{sec:asset}). For ISArt, rendered motion supplies evidence about
joint axes, pivots, ranges, and interference between parts. Unravel verifies
each removal edit and checks registration against attributed point clusters
and the observed extent of partially occluded objects
(Section~\ref{sec:scene}). Settling and visual audit assess the assembled
world. Packaging checks coordinate consistency, required files, and simulator
loadability. Each boundary returns a candidate and evidence, retaining failed
alternatives even when a later attempt succeeds.

\paragraph{Explicit execution and repair transitions.}
Each attempt separates observation, diagnosis, owner attribution, action,
validation, candidate selection, and routing. The transition contract is
formalized in Eq.~\eqref{eq:harness-transition}. The selected module returns a revised candidate
for validation; the decision advances execution, retries within budget,
restores a checkpoint, or records terminal failure. Visual critique can guide
refinement but cannot override a failed required geometric, physical, task,
or export check.


\paragraph{Evidence-driven owner selection.}
An unsuitable mesh returns to acquisition or geometry generation; an
incorrect joint returns to ISArt binding or structural planning. An
inconsistent removal plan or edit returns to the affected Unravel round.
Point attribution or alignment errors return to reconstruction or
registration, while penetration or unstable placement returns to recomposition
and settling. Packaging errors remain with export unless evidence identifies
an upstream cause. Records name affected entities, violated requirements,
supporting artifacts, and proposed owners. Ambiguous causes trigger further
inspection and remain explicit rather than receiving an unsupported diagnosis.

\paragraph{Repair scope and dependent-state invalidation.}
A repair affects an asset, part, instance, image region, removal round, point
cluster, or local placement workset. Replacing a mesh invalidates its
registration and physical checks on dependent instances. Revising an early
image edit can alter later removal images and fused point evidence, requiring
the affected downstream rounds to be reconstructed and checked again.
A pose change requires renewed collision, settling, and task-clearance checks
for affected objects. Other state is preserved only when its inputs remain
valid. Nested dispatch applies this rule
within specialized modules and across their boundaries.


\paragraph{Checkpoints and bounded recovery.}
The last validated checkpoint is retained before repair. A candidate is
selected only after required checks pass; improved appearance cannot
compensate for a broken joint, penetration, or missing task object.
Failed attempts consume the budget for their scope and may restore the
checkpoint. Exhaustion terminates the affected request with failure evidence.
Optional-object removal or another relaxation is recorded as degradation and
cannot satisfy an unchanged requirement. These mechanisms describe recovery
within a run rather than weight updates or guaranteed cross-run improvement.

\paragraph{World acceptance and demonstration admission.}
A world is accepted after its applicable semantic, geometric, physical, and
task requirements pass and a valid simulator package is produced
(Appendix~\ref{app:strict-predicate}). Scene-only requests produce reconstructed
scenes; promotion to an Accepted World additionally requires a task contract
and successful task validation. PolicyForge binds the accepted
world to an embodiment and reset specification, executes a teacher, and records
observations, states, and actions. World acceptance and episode success are
separate: a valid world can still yield a failed controller rollout.
Only episodes satisfying the task-success predicate enter the dataset.

\paragraph{Feedback, provenance, and reuse.}
PolicyForge returns world and task identities, the seed, affected entities,
and first blocker with a failed rollout
(Section~\ref{sec:policyforge-feedback}). Missing affordances, invalid geometry,
or world-owned physical failures request repair of the corresponding world
module; controller-owned failures remain with PolicyForge. A world revision
invalidates dependent resets and rollouts and must pass the same acceptance
gates before fresh demonstrations are admitted. Asset sources, world versions,
robot adapters, resets, and augmentation configurations remain associated with
accepted episodes so scene and task cousins can be traced to their collection
conditions.

Appendix~\ref{app:harness} supplies extended state and transition definitions.
Section~\ref{sec:experiments} measures the utility of the co-training data;
it does not independently measure repair efficiency or the harness contribution.

\FloatBarrier

%% file: sections/07_policy_generation.tex
\section{Policy Generation via Automated Demonstration Synthesis}
\label{sec:policyforge}

Given a validated Awomo-SimDataEngine world and a task contract, we propose \textbf{PolicyForge}, which provides a fully automated pipeline that converts a user-level task specification into large-scale, multi-embodiment demonstration data ready for policy learning. The pipeline operates within NVIDIA Isaac Sim and is designed to be embodiment-agnostic: it supports a growing roster of robotic platforms including Franka, Piper dual-arm, Agibot~G1, and Agibot~G2, each interacting with diverse objects across procedurally generated scenes.


\begin{figure}[htbp]
  \centering
  \includegraphics[width=\linewidth]{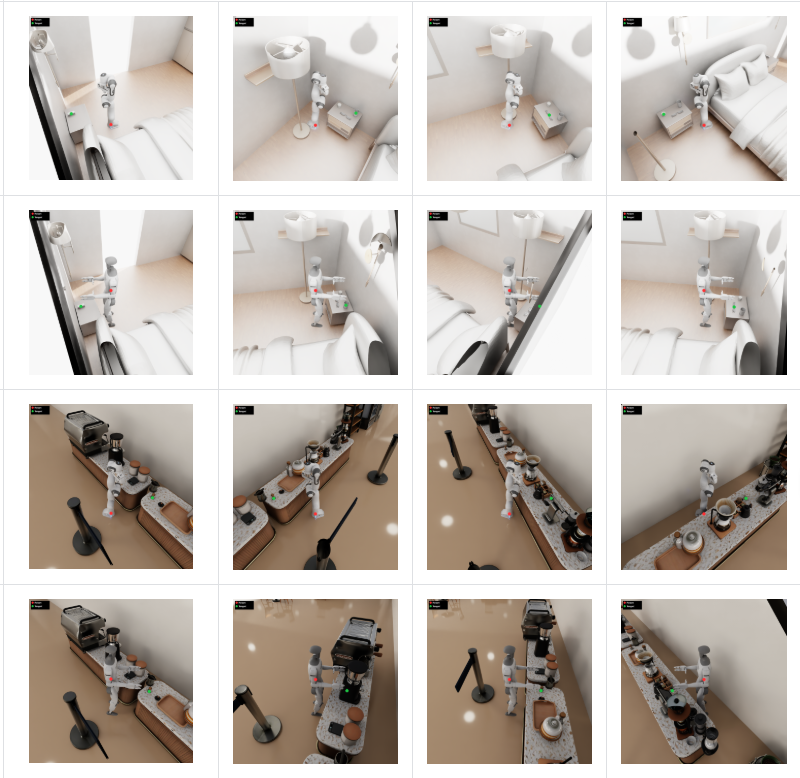}
  \caption{Qualitative examples of PolicyForge robot interactions
across different embodiments and scene configurations.
The $4 \times 4$ grid shows bedroom scenes in the top two rows
and coffee-shop scenes in the bottom two rows.
Examples illustrate variation in robot embodiment, object arrangement,
and viewpoint.}
  \label{fig:scene_embodiment}
\end{figure}

\subsection{Pipeline overview}
\label{sec:policyforge-overview}

\begin{figure}[htbp]
    \centering
    \includegraphics[width=\linewidth]{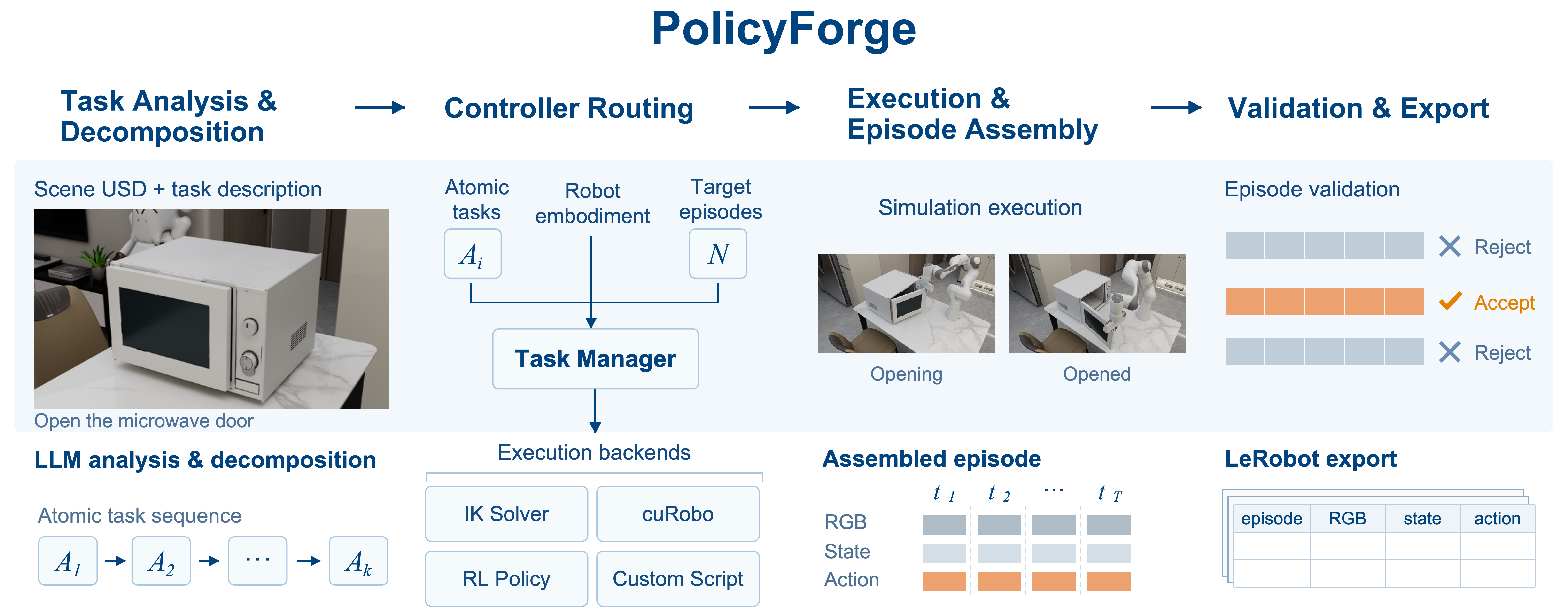}
    \caption{\textbf{PolicyForge pipeline.}
    Given a scene, task description, robot embodiment, and target episode count
    $N$, an LLM analyzes the task and decomposes it into atomic tasks when needed.
    A task manager selects suitable execution backends. Simulation execution in
    Isaac Sim produces synchronized observations, states, and actions, which are
    assembled into episodes. Episodes that pass task-success validation are
    exported in LeRobot format. The microwave images illustrate an ISArt input
    asset and its simulated manipulation.}
    \label{fig:policyforge_pipeline_v6}
\end{figure}

PolicyForge accepts three inputs: (i)~a validated scene generated through either Unravel or SimForge with retrieved, generated,
or procedurally modeled assets, (ii)~a task description specifying
goal predicates, interaction requirements, and success/termination conditions,
and (iii)~an embodiment record that fixes the robot model, sensor suite, and
action parameterization. A target episode count $N$ specifies the requested
generation scale. Figure~\ref{fig:policyforge_pipeline_v6} summarizes four stages:

\begin{enumerate}
  \item \textbf{Task analysis and decomposition.} An LLM analyzes the scene and
        task description and, when needed, decomposes the task into an ordered
        sequence of atomic tasks $A_1,\ldots,A_k$. An already atomic task is
        represented by a single element ($k=1$).
  \item \textbf{Controller routing.} A task manager selects an execution backend
        for each atomic task according to the task requirements and robot
        embodiment. Available backends include an inverse-kinematics solver,
        cuRobo motion planning, an RL policy, and custom scripts. The selected
        backend is specific to the data-generation configuration.
  \item \textbf{Execution and episode assembly.} The scene graph is instantiated
        in Isaac Sim with physics-consistent placements, support/containment
        relationships, and randomized nuisance parameters (lighting, textures,
        object poses). A reset controller samples a robot and object configuration
        satisfying the task preconditions and records the simulator state for
        replay. The selected backends execute the atomic-task sequence while
        recording multi-camera RGB observations, proprioceptive state, and
        end-effector actions at the control rate. These synchronized streams are
        assembled into an episode.
  \item \textbf{Validation and export.} Each episode is validated against the
        task-success predicate; only successful episodes are retained. The
        accepted episodes are exported in LeRobot format with aligned image,
        state, and action streams for downstream policy learning.
\end{enumerate}

\subsection{Task contract}
\label{sec:policyforge-task-contract}

The pipeline consumes a validated tuple
\[
  (\mathcal{W},\;\mathcal{T},\;R_{\mathrm{bot}},\;\Sigma),
\]
where $\mathcal{W}$ is the compiled scene, $\mathcal{T}$ encodes the task goal,
entities, interaction requirements, and success/termination predicates,
$R_{\mathrm{bot}}$ specifies the embodiment and observation/action interface,
and $\Sigma$ specifies reset state, randomization ranges, episode horizon, and
seed.  The contract is expressed over exported object, support, container, and
articulation identities rather than hidden generator internals, ensuring full
reproducibility and downstream verifiability.

\subsection{Multi-embodiment adaptation}
\label{sec:policyforge-embodiment}

A key design principle of PolicyForge is embodiment decoupling: the scene and
task specification are independent of the robot.  An embodiment adapter binds
kinematics, sensors, gripper type, control rate, and action parameterization
to the shared world and task record.  This enables the same scene--task pair to
generate demonstrations for heterogeneous platforms---from industrial Franka arm setups to humanoid-scale systems---without modifying the
upstream scene graph.  The adapter also normalizes action and proprioceptive
representations so that data produced for different embodiments can be mixed at
training time after a lightweight convention alignment step.

\subsection{Failure feedback}
\label{sec:policyforge-feedback}

The failure schema classifies rollout failures into six categories---task
specification, missing affordance, geometry/support, physics, reset, and
controller---when the registered manifest supplies sufficient evidence.
A world-owned failure feeds back the affected entities, first blocker, and a
requested change to the smallest responsible graph edge to the Awomo-SimDataEngine harness
for automated scene repair.  Every requested revision must pass the same
validation and strict export gates before contributing new rollouts,
preventing regression loops.

\subsection{Data cousins: scaling diversity along scene and task axes}
\label{sec:policyforge-cousins}

Inspired by the digital-cousin paradigm, PolicyForge systematically generates
data variants---\emph{cousins}---along two orthogonal axes:

\begin{itemize}
  \item \textbf{Scene cousins.}  The same task executed in visually and
        physically distinct environments (e.g., kitchen $\to$ bedroom $\to$
        workshop), varying backgrounds, lighting, and distractor objects.
  \item \textbf{Task cousins.}  Variations of the manipulation goal within the
        same scene (e.g., \emph{pick mug} $\to$ \emph{pick bowl} $\to$
        \emph{stack plates}), diversifying the interaction semantics while
        sharing the visual context.
\end{itemize}

This cousin-based decomposition is the basis for the data families evaluated
in Section~\ref{sec:experiments}: \textsc{SimData-Easy} (straightforward
scenes and tasks), \textsc{SimData-Lite} (simple scenes with challenging
manipulation tasks), and \textsc{SimData-Hard} (complex scenes paired with
demanding task specifications), each representing a distinct point in the
scene$\times$task difficulty space. The \textsc{SimData-Ultra} co-training
configuration additionally mixes data sources and applies robot initial-state
perturbations and language-instruction augmentation, as specified in
Section~\ref{sec:exp-setup}.

\section{Experiments and Results}
\label{sec:experiments}

We evaluate whether PolicyForge demonstrations generated in Isaac Sim improve
a World-Action Model (WAM) on the MuJoCo-based LIBERO-Plus benchmark~\citep{fei2025liberoplus}.
The main comparison selects each configuration's best evaluated checkpoint
up to and including 13K steps by overall success, using the same evaluation
set of 1,003 episodes across object, spatial, goal, and long-horizon suites. This measures the downstream utility of the co-training data across
simulator boundaries. Detailed checkpoint results and additional
LIBERO-Pro~\citep{zhou2025liberopro} evaluations are provided in Appendix~\ref{sec:appendix-data}.

\subsection{Setup}
\label{sec:exp-setup}

\paragraph{Policy and training protocol.}
WAM uses a video and action diffusion-transformer architecture initialized
from Wan2.2-TI2V-5B~\citep{wan2025}. Inputs comprise front and wrist RGB views plus
proprioception; outputs encode end-effector motion and gripper state.
LIBERO-Plus experiments use 32 GPUs, global batch size 768, and the training
recipe in Appendix~\ref{sec:appendix-training-hp}. The baseline uses only the
benchmark demonstrations. Co-training configurations use weighted sampling
with a reported 1:1 effective-frame ratio between benchmark and additional
data. All headline comparisons use the best evaluated checkpoint up to and including
13K steps for each configuration. Selection uses overall success on the
reported evaluation set; this is a best-observed comparison, not selection
on a separate validation set. Suite scores come from the same selected
checkpoint rather than separate per-suite maxima.

\paragraph{Data configurations.}
\textsc{SimData-Easy} uses straightforward scenes and pick-and-place tasks;
\textsc{SimData-Lite} combines simpler scenes with more demanding
manipulation tasks; \textsc{SimData-Hard} combines richer scenes and task
specifications. \textsc{SimData-Ultra} mixes \textsc{SimData-Hard}, additional
in-house demonstrations from complex scenes, and the benchmark demonstration
source, with robot initial-state perturbations and
language-instruction augmentation. It is a combined data configuration:
the experiment does not isolate these interventions individually.
Data-source records and sampling configuration are given in
Appendix~\ref{sec:appendix-data-config}.

\subsection{Cross-Simulator Co-training Improves Overall Success}
\label{sec:exp-cross-sim}

\begin{figure}[htbp]
\centering
\includegraphics[width=0.9\linewidth]{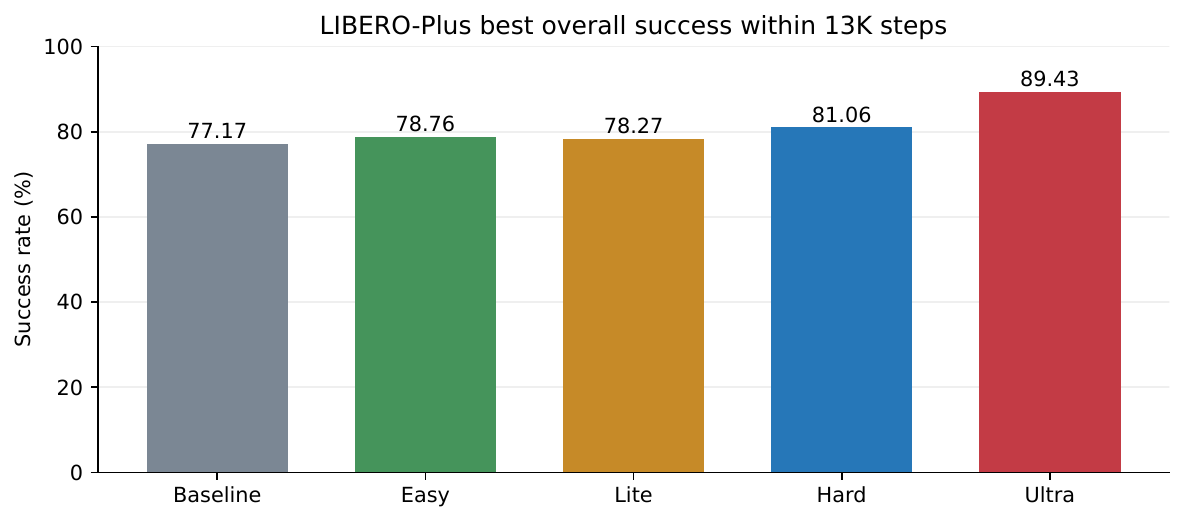}
\caption{\textbf{Best-checkpoint co-training comparison.} Best observed overall
WAM success on LIBERO-Plus up to and including 13K steps. SimData-Ultra gives the highest score.}
\label{fig:exp-plus-overall}
\end{figure}

Among evaluated checkpoints up to and including 13K steps,
\textsc{SimData-Ultra} achieves its best overall success of
\textbf{89.43\%} at 13K, compared with the baseline's best
\textbf{77.17\%} at 9K:
an improvement of \textbf{12.26 percentage points}. It also exceeds
\textsc{SimData-Hard} by \textbf{8.37 points}. All three other Awomo-SimDataEngine
configurations improve on the baseline under this selection rule
(Table~\ref{tab:exp-plus-main} and Figure~\ref{fig:exp-plus-overall}).

\begin{table}[htbp]
\centering
\small
\caption{WAM success on LIBERO-Plus at each configuration's best evaluated
checkpoint up to and including 13K steps, selected by overall success.
The gain is measured against benchmark-only training in percentage points.
All configurations use the same 1,003-episode evaluation set.}
\label{tab:exp-plus-main}
\begin{tabular}{lrrr}
\toprule
Co-training data & Selected step & Success (\%) & Gain (pp) \\
\midrule
Baseline & 9K & 77.17 & --- \\
SimData-Easy & 13K & 78.76 & +1.59 \\
SimData-Lite & 13K & 78.27 & +1.10 \\
SimData-Hard & 13K & 81.06 & +3.89 \\
\textbf{SimData-Ultra} & 13K & \textbf{89.43} & \textbf{+12.26} \\
\bottomrule
\end{tabular}
\end{table}

\subsection{Goal Generalization Shows the Largest Gain}
\label{sec:exp-diversity}

\begin{figure}[htbp]
\centering
\includegraphics[width=\linewidth]{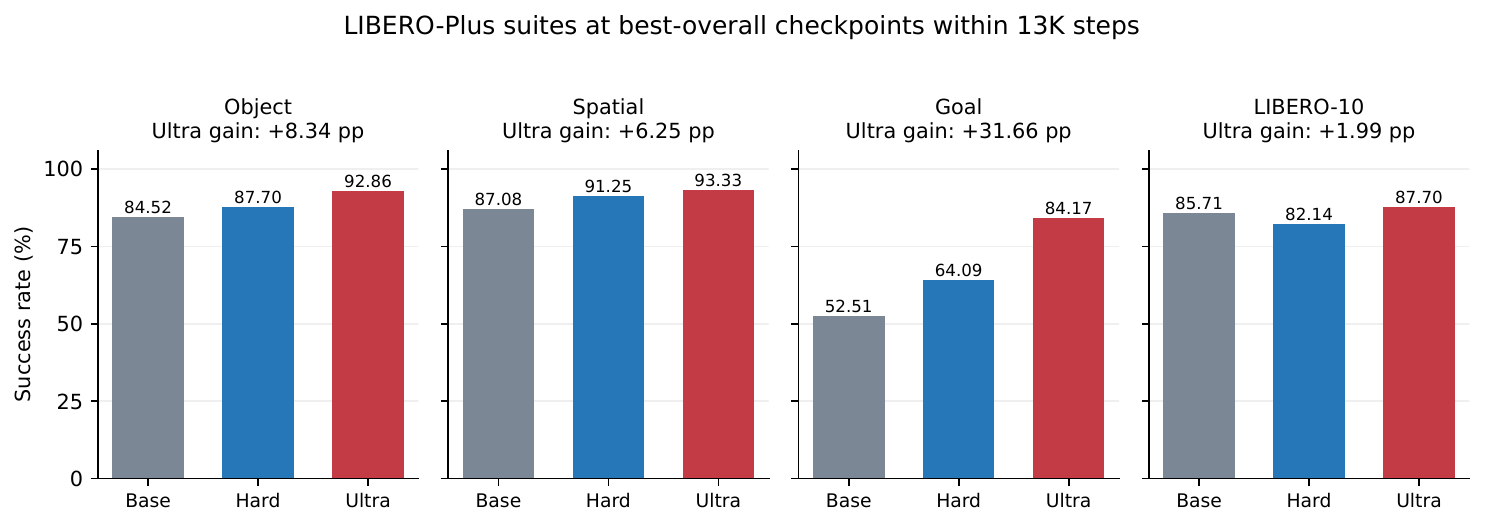}
\caption{\textbf{Generalization at the selected best-overall checkpoints.}
Each configuration is selected by its highest overall success up to and
including 13K steps; all four suites use that selected checkpoint. Gains are largest for goals,
followed by objects and spatial configurations; the long-horizon improvement is limited.}
\label{fig:exp-plus-profile}
\end{figure}

Figure~\ref{fig:exp-plus-profile} compares the baseline, Hard, and Ultra at
their respective best-overall checkpoints up to and including 13K steps
(9K, 13K, and 13K, respectively). With \textsc{SimData-Ultra}, goal generalization
increases from \textbf{52.51\%} to \textbf{84.17\%}
(\textbf{+31.66 points}), spatial generalization from \textbf{87.08\%} to
\textbf{93.33\%} (\textbf{+6.25 points}), and object generalization from
\textbf{84.52\%} to \textbf{92.86\%} (\textbf{+8.34 points}).
Relative to Hard, Ultra adds \textbf{20.08 points} on the goal suite.
The pattern is consistent with a richer training distribution across
objects, layouts, goals, initial states, and instruction wording. Establishing
the independent contribution of each factor requires controlled ablations.

Long-horizon LIBERO-10 improves from $85.71\%$ to $87.70\%$
($+1.99$ points), a much smaller change. The synthetic data provides limited
long-horizon sequential supervision, so the result supports stronger object,
spatial, and goal generalization rather than a broad claim that sequential
manipulation has been solved.

\subsection{Training Progress and Best-Checkpoint Performance}
\label{sec:exp-diversity-engine}

\begin{figure}[htbp]
\centering
\includegraphics[width=0.95\linewidth]{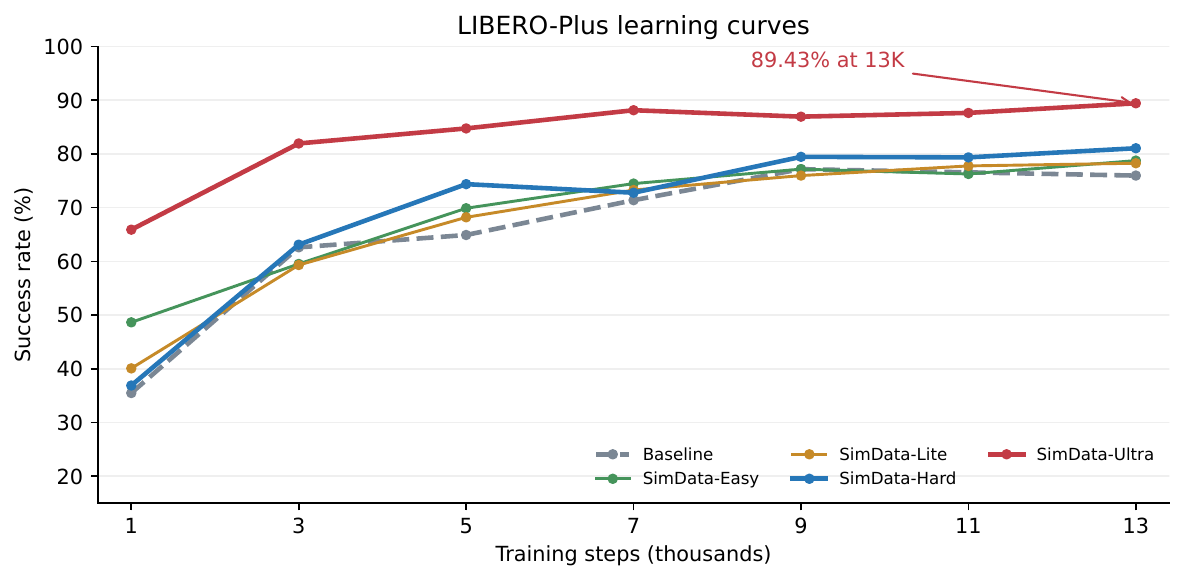}
\caption{\textbf{WAM learning curves on LIBERO-Plus.} Points correspond to
the reported evaluated checkpoints up to and including 13K steps. Ultra
reaches $84.75\%$ at 5K and its best score within this budget is
$89.43\%$ at 13K. Main comparisons select each configuration's
highest-overall-success checkpoint within the same budget.}
\label{fig:exp-plus-curves}
\end{figure}

The learning curves in Figure~\ref{fig:exp-plus-curves} show that Ultra
reaches $84.75\%$ at 5K steps, exceeding Hard's highest listed score
($81.06\%$ at 13K). Within the 13K-step selection budget, Ultra's
best evaluated checkpoint reaches \textbf{89.43\%} at 13K, whereas the
baseline's best score is \textbf{77.17\%} at 9K, an
\textbf{12.26-point} difference. Later checkpoints are retained in the
appendix but excluded from the main comparison. The curves indicate earlier attainment of high success;
they do not measure wall-clock savings or data-generation cost.

\subsection{Summary and Scope}
\label{sec:exp-summary}

The main LIBERO-Plus comparison uses the best-overall evaluated checkpoints
up to and including 13K steps:
\textbf{+12.26 points overall}, \textbf{+31.66 on goals}, and
\textbf{+6.25 on spatial generalization}. These measurements support the
utility of the integrated PolicyForge co-training data across simulator
boundaries. Additional LIBERO-Pro results are provided in the appendix.
The reported tables do not include repeated-training-seed confidence
intervals, and the combined Ultra configuration is not an ablation of ISArt,
Unravel, the harness, or individual augmentations. Physical-world transfer
and world-generation acceptance rates require separate evaluation.

%% file: sections/09_conclusion.tex
\section{Limitations and Conclusion}
\label{sec:limitations}

\subsection{Limitations}

\paragraph{Single-view ambiguity and structural priors.}
ISArt and Unravel infer geometry and interaction structure from incomplete
visual evidence. ISArt's part geometry depends on predicted semantics,
boxes, cavities, and kinematic relations; incorrect priors can require
structural re-planning or geometry regeneration. Unravel's image edits
hypothesize occluded surfaces, and fused point maps inherit uncertainty from
both image completion and monocular reconstruction. Motion validation,
registration, and settling check candidate consistency but cannot establish
that hidden geometry or metric scale matches the physical scene. The current
asset and physics representations primarily support rigid objects and
articulated mechanisms; deformable interaction requires additional modeling.

\paragraph{Validation and bounded recovery.}
The Graph-Native Harness makes evidence, dependencies, repair scopes, and
checkpoint decisions explicit. Its validators and critics can still miss
errors or attribute them incorrectly, and a finite repair budget can end
with an unresolved request. World acceptance is conditional on the applicable
checks and simulator contract. It does not guarantee successful execution of
every downstream task. The reported experiments do not quantify harness
repair efficiency, generation cost, or end-to-end accepted-world yield.

\paragraph{Demonstration coverage and sequential tasks.}
PolicyForge depends on scripted or motion-planned teachers, feasible resets,
and task-success predicates. Demonstrations therefore cover interactions
that these teachers can execute and validate. Contact-rich manipulation,
dexterous control, and extended task sequences remain challenging.
On LIBERO-Plus, Ultra's gain on LIBERO-10 is $1.99$ percentage points at
each configuration's best-overall evaluated checkpoint up to and including
13K steps, compared with $31.66$ on goals and $6.25$ on spatial generalization.
This limited sequential-task gain motivates richer task composition and
long-horizon demonstration coverage; the current comparison does not isolate
which data factors explain the difference.

\paragraph{Experimental scope and attribution.}
The module and policy evaluations answer different questions.
ISArt's quantitative binding experiment supplies ground-truth part geometry,
so its reported binding success does not measure end-to-end asset generation.
Unravel's five-scene coverage comparison uses its own masks and point clusters
as references; increased coverage and point count do not establish accuracy
against independent geometric ground truth. Its asset-fidelity scores average
160 objects, without reported confidence intervals or rater-agreement
statistics. The WAM experiments evaluate the combined Ultra data
configuration, without repeated-training-seed confidence intervals or
controlled ablations of ISArt, Unravel, harness repair, and individual
augmentations. The WAM evaluations show that gains vary between
LIBERO-Plus and LIBERO-Pro. Evaluation across Isaac Sim and MuJoCo supports
cross-simulator training utility; real-robot transfer remains unmeasured.

\subsection{Conclusion}
\label{sec:conclusion}

Awomo-SimDataEngine connects interactive assets, text- and image-driven scene generation, and
robot demonstration synthesis through a shared Graph-Native Harness.
ISArt couples geometric--kinematic planning with structure-conditioned part
generation, articulation binding, and simulation feedback. Unravel links
verified deconstruction to point-map fusion, object registration,
reverse-order recomposition, and physical settling. SimForge constructs
single-room and multi-room scenes from text. The harness coordinates
these modules through typed state, validation evidence, bounded repair, and
checkpoints. PolicyForge binds accepted worlds to tasks, embodiments, and
replayable resets, then admits successful teacher rollouts as demonstrations.
Together these components provide a traceable asset--world--demonstration
workflow for expanding robot-training data.

The experiments support this workflow at complementary levels.
On the 60-object geometry benchmark, ISArt achieves CD $0.163$, compared
with $0.268$ for TRELLIS.2 + P3-SAM + X-Part.
Unravel improves blind human asset-fidelity scores from $4.9$ to $8.6$
over 160 objects. For downstream learning, Isaac Sim demonstrations improve
WAM success on MuJoCo-based LIBERO-Plus from $77.17\%$ to $89.43\%$
at each configuration's best evaluated checkpoint up to and including 13K
steps, selected by overall success. This is a $12.26$-point overall gain,
with improvements of $31.66$ points on goals and $6.25$ on spatial
generalization. These results establish module-level
geometry and fidelity improvements and the utility of the combined
co-training data under their respective protocols. Priorities for further
work include independent scene-geometry evaluation, measured end-to-end
readiness and repair costs, controlled module and augmentation ablations,
and broader sequential-task and real-robot evaluation.

%% file: sections/appendix_harness.tex
\section{Graph-Native Harness}
\label{app:harness}

\paragraph{World state.}
The constructed world is represented by
\begin{equation}
S=(R,O,A,X,H,J,P,M).
\end{equation}
$R$ denotes rooms, architectural structure, and opening geometry. $O$ denotes object instances with stable scene identities. $A$ denotes visual, collision, and articulated assets. $X$ stores world poses for objects and reference poses for articulated links. $H$ stores support, containment, room ownership, and parent lineage. $J$ stores joints, types, axes, limits, dynamics, and serialized articulation state. $P$ stores mass, friction, inertia, stability, and collision semantics. $M$ stores materials, textures, and appearance attributes.

The world state is intentionally separate from evidence. The same $S$ may be retained, repaired, or rejected under different validation histories, so the evidence plane must be explicit rather than implicit.

\paragraph{Evidence state.}
The evidence state is
\begin{equation}
E=(E_{\mathrm{sem}},E_{\mathrm{geo}},E_{\mathrm{phy}},E_{\mathrm{vis}},E_{\mathrm{exp}}).
\end{equation}
The components denote semantic, geometric, physical, visual, and export evidence. A hard failure in any of these components can block promotion even if the rendered scene looks plausible.

\paragraph{Harness state.}
The execution state at step $k$ is
\begin{equation}
H_k=(S_k,T,E_k,R_k,B_k,\kappa_k),
\end{equation}
where $R_k$ is the selected graph edge and causal owner, $B_k$ is the remaining retry
budget, and $\kappa_k$ is the checkpoint selected for preservation or rollback.
This is the normative Harness state used to define the transition semantics;
module implementations retain stage snapshots and artifact records; the schema
specifies their common audit meaning rather than a universal serialization.

\paragraph{Accepted-world gate.}
A world is simulation-ready only when the compiled task and the constructed world satisfy the evidence gate
\begin{equation}
\Phi(S,T,E)=\mathbf{1}[\text{semantic} \wedge \text{geometric} \wedge \text{physical} \wedge \text{task} \wedge \text{export}].
\end{equation}
We call the result an \emph{Accepted World} exactly when $\Phi(S,T,E)=1$.
The typed request enters the Harness at the input boundary. Scene-only
requests return reconstructed scenes without Accepted World status;
task-conditioned requests return an Accepted World only after the task
contract and all required checks pass. Downstream policy evaluation and
demonstration synthesis consume only Accepted Worlds. A saved render, partial
layout, or best-effort export is therefore not an accepted-world output.
Module-specific image reconstruction and demonstration synthesis are described
in Sections~\ref{sec:scene} and~\ref{sec:policyforge} respectively.

\subsection{Transition and Repair Semantics}
\label{sec:quality}

Awomo-SimDataEngine treats iterative generation as an explicit harness, not as a hidden
retry loop. The graph does not merely host stages; it atomizes adaptation into
typed transitions, evidence-bearing ownership, and bounded re-entry.
The transition is the method's smallest unit: models, critics, solvers, and
simulators are executors attached to it, not competing control abstractions.

\subsection{Control Plane}
The control plane is the role of the orchestrator.  It inspects typed state and
structured evidence, applies a bounded budget, and chooses the owner and next
edge.  An executor may be learned, deterministic, or tool-backed; the control
contract remains the same.

\begin{table}[t]
\centering
\small
\begin{tabularx}{\linewidth}{>{\RaggedRight\arraybackslash}p{0.24\linewidth} >{\RaggedRight\arraybackslash}p{0.16\linewidth} >{\RaggedRight\arraybackslash}p{0.28\linewidth} >{\RaggedRight\arraybackslash}X}
\toprule
Observation & Owner & Repair scope & Evidence \\
\midrule
Incomplete or inconsistent object-removal plan or image edit & Unravel Planning / Editing & Affected removal round and image region & Plan, mask, and edit-verification evidence \\
Missing, rejected, or semantically wrong asset & Asset & Named request and dependent scene instances & Acquisition and semantic evidence \\
Mismatch between an asset and its observed point cluster & Registration & Named object and attributed point cluster & Point-cloud, mask, and alignment evidence \\
Residual penetration, physical instability, or simulator infeasibility & Physics / Settling & Affected object instances and local recomposition workset & Physical feasibility evidence \\
Validation regression without a hard geometry failure & Critique & Structured modification plan with a bounded attempt budget & Perceptual and validation evidence \\
Missing or invalid simulator package, frame, or strict-export predicate & Export / Packaging & Package manifest and coordinate/export artifacts; return upstream only when evidence identifies a source-geometry or physics cause & Packaging evidence \\
\bottomrule
\end{tabularx}
\caption{Owner-aware repair contract.  The repair scope follows the causal
owner when it can be identified; unrelated validated state remains protected by
the checkpoint policy.}
\label{tab:repair-routing}
\end{table}

Structured evidence names an owner, a bounded scope, and the checkpoint
decision needed before the next transition.

\subsection{Loop-to-Graph Atomicization}
The point of the Harness is not that the system has a loop.  The point is that
the loop has been decomposed into first-class transitions:
\begin{equation}
\label{eq:harness-transition}
\begin{aligned}
o_k=\operatorname{Observe}(H_k),\quad
d_k=\operatorname{Diagnose}(o_k),\quad
u_k=\operatorname{Attribute}(d_k,H_k),\quad
a_k=\operatorname{Act}_{u_k}(H_k),
\\[-2pt]
v_k=\operatorname{Validate}(a_k),\quad
H_{k+1}=\operatorname{Select}(H_k,a_k,v_k),\quad
r_{k+1}=\operatorname{Route}(v_k,H_{k+1}).
\end{aligned}
\end{equation}
Here $u_k$ is the causal owner, $H_{k+1}$ is the selected candidate or
checkpointed state, and $r_{k+1}$ is the next explicit graph edge.  The atomic
transition therefore carries typed state, structured evidence, repair scope, a
retry budget, and a checkpoint decision.  This is the single control semantics
used by every domain subgraph.

\subsection{Evidence Routing}
The transition semantics above are evidence-driven, not score-driven.  Visual,
geometric, physical, and packaging observations are normalized into a common
evidence record before an owner is selected.  When the cause is ambiguous, the
Harness preserves the ambiguity and checkpoint rather than inventing a precise
owner; unresolved cases remain explicit non-successes.

\subsection{Scoped Re-entry}
Re-entry returns to the smallest owner that can change the failing state.  The
owner table supplies the canonical mapping, and unaffected validated state
remains eligible for reuse while the repair consumes only the remaining budget
for that scope.

\subsection{Checkpoint, Rollback, and Degradation}
Checkpoints are not decorative saves.  A candidate is selected only after
validation; if it improves one metric but violates a hard predicate, the
Harness restores the last valid checkpoint or records the actual degradation
instead of pretending the run succeeded.

Degradation records capture observed execution failures, never success.  They
remain part of the audit trail and cannot substitute for validation.

\subsection{Nested Subgraphs}
The global Harness composes owner-local subgraphs around the world compiler
described in Section~\ref{sec:representation}.  Each subgraph may contain local
transitions, but all of them use the single contract above.  Strict export
closes a local validation attempt by promoting only an accepted candidate;
multimodal adapters and policy interfaces connect at that boundary without
introducing additional control mechanisms.

\subsection{Multimodal Condition and Typed Specification}
\label{app:condition-schema}

The condition contract makes modality an input adapter, not a separate control
logic. SimForge uses text requests directly; Unravel uses a single scene image
with an optional description. Object-centric references
provide the separate ISArt input.  The table lists conceptual groups rather than
a prescribed serialization format.

\begin{longtable}{@{}p{0.24\linewidth}p{0.68\linewidth}@{}}
\caption{Conceptual condition and typed-specification contract.}\label{tab:condition-schema}\\
\toprule
Group & Required meaning \\
\midrule
\endfirsthead
\toprule
Group & Required meaning \\
\midrule
\endhead
Condition identity & Stable cohort identifier, modality label, source provenance, calibration status, and adapter/version owner. \\
Scene request & Rooms, architectural elements, object requests, attributes, dimensions, relations, and material/style constraints. \\
Task request & Goal, ordered subgoals, operable entities, interaction type, robot constraints, and success predicates. \\
Compilation evidence & Conflicts between modalities, unresolved requirements, confidence, and the rule used to resolve each conflict. \\
Construction requirements & Asset, articulation, support, clearance, containment, and export prerequisites derived from the task. \\
Output contract & Typed scene--task specification consumed by the ordered
planning, acquisition, construction, validation, and export stages; an
optional policy handoff attaches only after the Accepted World. \\
\bottomrule
\end{longtable}

\subsection{Harness State Fields (implementation: PipelineState)}
\label{app:pipeline-state}

\texttt{PipelineState} is the implementation name for the Harness state
exchanged by graph nodes.  The state is defined by semantic responsibilities,
not by a public serialization shape.
Snapshots may be partial during diagnosis, but a partial snapshot is never a
strict success claim.

\begin{longtable}{@{}p{0.27\linewidth}p{0.65\linewidth}@{}}
\caption{Conceptual Harness-state field groups (implementation name:
\texttt{PipelineState}).}\label{tab:pipeline-state}\\
\toprule
Field group & Required audit meaning \\
\midrule
\endfirsthead
\toprule
Field group & Required audit meaning \\
\midrule
\endhead
Condition and configuration & Source identity, modality, adapter/version, model choices, seeds, strictness, and declared budgets. \\
Scene and layout & Current typed scene state, room topology, relation diagnostics, and the stage at which the snapshot was taken. \\
Acquisition & Asset requests, candidate outcomes, accepted sources, rejection reasons, and terminal failures keyed by request identity. \\
Placement and physics & Poses, support relations, collision and stability evidence, hard violations, and object removals or restorations. \\
Critique and evidence & Structured visual, geometric, physical, and export observations with severity, entities, and candidate owner. \\
Control and recovery & Selected graph edge, retry budget, checkpoint choice, rollback history, regression status, and next edge. \\
Degradation and provenance & Observed execution outcome, unresolved limitations, responsible owner, and versioned provenance.  Degradation is never success evidence. \\
\bottomrule
\end{longtable}

\subsection{SceneState and Object Fields}
\label{app:scene-state}

\texttt{SceneState} is the candidate-world checkpoint.  It distinguishes an
asset template from a scene instance and preserves the lineage needed for
support-aware repair and strict export.

\begin{longtable}{@{}p{0.27\linewidth}p{0.65\linewidth}@{}}
\caption{Conceptual \texttt{SceneState} and object-instance contract.}\label{tab:scene-state}\\
\toprule
Group & Required audit meaning \\
\midrule
\endfirsthead
\toprule
Group & Required audit meaning \\
\midrule
\endhead
World identity & Stable scene identity, condition linkage, stage, iteration, and checkpoint provenance. \\
Topology & Rooms, boundaries, openings, architecture, object membership, and typed spatial relations. \\
Object instance & Unique instance identity, semantic role, placement role, room ownership, dimensions, pose, and measured geometry. \\
Asset evidence & Non-placeholder visual and collision representations, material/physical properties, source provenance, and validation status. \\
Support lineage & Parent instance, concrete support surface, local pose, footprint, clearance, and sibling compatibility for every supported child. \\
Articulation & Links, joint type, axis, origin, limits, and actuation metadata for movable objects. \\
Constraints and history & Requested relations, repair markers, validator outcomes, and prior accepted checkpoint references. \\
\bottomrule
\end{longtable}

\subsection{Strict Export Predicate}
\label{app:strict-predicate}

The report uses a conjunctive export gate:
\begin{equation}
\begin{aligned}
\mathsf{StrictSuccess}\iff{}&\text{terminal export stage}\land\text{successful execution}\\
&\land\text{hard validation passes}\land\text{validated simulator package}.
\end{aligned}
\end{equation}
File existence, a render, or an intermediate package cannot establish any
conjunct.  Accepted-scene full success additionally requires experiment-level
loadability, visual/collision alignment, task-entity coverage, and reset
replay.  Unresolved degradation, incomplete execution, and best-effort output
remain diagnostic failures.

\subsection{Evidence Package Organization}
\label{app:artifact-layout}

The evidence plane is organized by meaning rather than by storage location.
Later layers may be absent when an earlier stage fails; absence is recorded as
a failure state, not treated as an empty success.

\begin{table}[t]
\centering
\small
\caption{Conceptual organization of the evidence package.}
\label{tab:evidence-package}
\begin{tabularx}{\linewidth}{@{}p{0.23\linewidth}X p{0.22\linewidth}@{}}
\toprule
Layer & Contents & Strictness role \\
\midrule
Condition & Modality, source identity, adapter/version, task, seeds, and split & Immutable cohort key \\
Scene state & Rooms, objects, poses, relations, assets, joints, support lineage, and materials & Candidate world state \\
Execution trace & Transitions, observations, owner decisions, budgets, checkpoints, and rollbacks & Harness audit trail \\
Asset evidence & Candidate sources, geometry, collision, material, and validation outcomes & Acquisition gate \\
Validation evidence & Semantic, visual, geometric, physical, export, and degradation records & Hard/soft gate inputs \\
Simulator package & Coordinate contract, dependencies, load result, and reset metadata & Strict export gate \\
Policy handoff & Accepted world, task contract, robot/reset adapter, rollout status, and first blocker & Downstream interface \\
\bottomrule
\end{tabularx}
\end{table}

\subsection{Per-Run Evidence Checklist}
\label{app:evidence-checklist}

Before a run contributes to accepted-scene or policy-facing metrics, the
following conceptual evidence must be present:

\begin{enumerate}[leftmargin=*,nosep]
\item A condition record binds modality, source identity, adapter/version, task,
seeds, split, and protocol version.
\item The final state records stage, warnings, errors, transitions, checkpoints,
rollback decisions, and every degradation with its actual owner and reason.
\item Rooms, objects, poses, dimensions, relations, and placement evidence are
present in the scene checkpoint.
\item Every accepted object has non-placeholder visual/collision evidence,
provenance, canonical geometry, and validation status.
\item Rejected and failed candidates remain in the negative-evidence ledger.
\item Every supported child has concrete parent/surface lineage, local pose,
measured hull, clearance, and sibling-collision evidence.
\item Geometry, physics, articulation, and export validators report hard
violations and first blockers; visual critique remains a soft observation.
\item The strict conjunctive export gate passes; a render, intermediate package,
or best-effort artifact alone never qualifies.
\item Policy episodes, when evaluated, join world, task, robot, simulator,
seed, checkpoint, terminal status, first blocker, and recovery evidence.
\item Environment and resource records distinguish generation, export, physics,
and policy costs and preserve the declared version/budget contract.
\end{enumerate}

%% file: sections/appendix_isart.tex
\section{ISArt}
\label{app:isart}

\subsection{Structural Encoding and Geometry Conditioning}
\begin{figure}[t]
\centering
\includegraphics[width=0.95\linewidth]{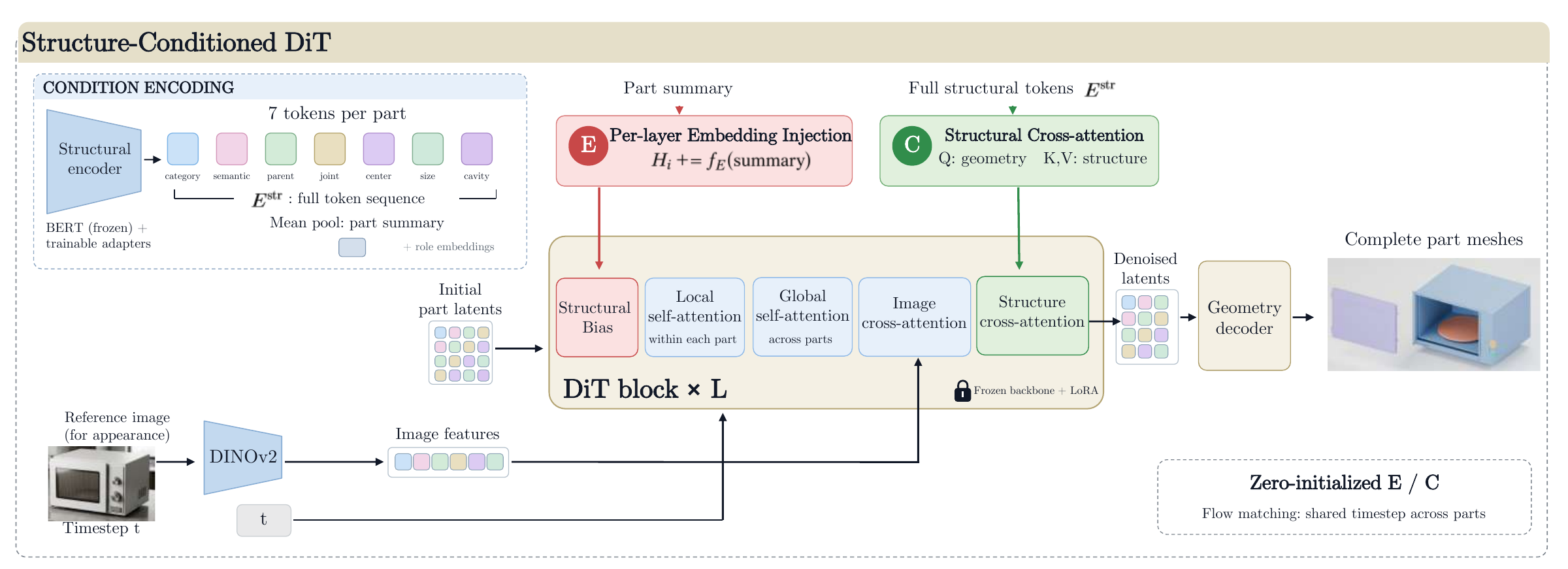}
\caption{ISArt's structure-grounded multi-part DiT. E and C condition the
pretrained geometry backbone while retaining visual conditioning.}
\label{fig:isart-dit}
\end{figure}

ISArt introduces a structure-conditioned DiT-based geometry generation framework that explicitly injects semantic, spatial, and kinematic priors into the multi-part denoising process. The overall architecture is illustrated in Fig.~\ref{fig:isart-dit}. To acquire the geometry and kinematic structure of the target object, we first use a VLM to predict the global structural information
\begin{equation}
\mathcal{S}
=
\left(
N, T_{\text{category}}
\right),
\end{equation}
where $N$ denotes the number of parts and
$T_{\text{category}}$ denotes the object category.
We then jointly predict structural information $\{S_i\}_{i=1}^{N}$ for all parts. For part $i$, we organize part-wise structural information into several text descriptions and vectors: 

\begin{equation}
S_i
=
\left(
T_{\text{semantic},i},
T_{\text{conn},i},
l_i,
J_i,
\mathbf{c}_{\text{box}, i},
\mathbf{s}_{\text{box}, i},
\mathbf{c}_{\text{cav}, i},
\mathbf{s}_{\text{cav}, i}
\right)
\end{equation}
where $T_{\text{semantic},i}$ denotes the semantic label of part $i$;
$T_{\text{conn},i}$ denotes its parent and child connection in the kinematic tree;
$l_{i} \in \{0,1\}^{4}$ denotes the one-hot joint-type representation of part $i$, corresponding to four predefined categories: fixed, revolute, prismatic, and continuous;
$J_i \in \mathbb{R}^3$ denotes the coarse joint-axis direction of part $i$;
$\mathbf{c}_{\text{box},i}, \mathbf{s}_{\text{box},i} \in \mathbb{R}^3$
denote the center and size of the part bounding box, and $\mathbf{c}_{\text{cav},i}, \mathbf{s}_{\text{cav},i} \in \mathbb{R}^3$ denote the center and size of the associated cavity bounding box. The proposed method only handles the case of one predicted cavity (one hole). If no cavity is predicted, we set $\mathbf{c}_{\text{cav},i} = \mathbf{s}_{\text{cav},i} = \mathbf{0}$. We encode the textual attributes
$\{T_{\text{semantic},i},T_{\text{conn},i}\}$
using a frozen BERT-base encoder followed by a learnable projection to the structural feature dimension $d_s$, while the discrete joint-type attribute $l_i$ is mapped through a learned embedding.
The continuous canonical-frame attributes
$\{J_i,\mathbf{c}_{\text{box},i},\mathbf{s}_{\text{box},i},
\mathbf{c}_{\text{cav},i},\mathbf{s}_{\text{cav},i}\}$
are independently projected by lightweight MLPs into the same $d_s$-dimensional space; attribute-specific role embeddings are added to distinguish their semantics, and learned missing-value embeddings are used whenever an attribute is unavailable.

Given the structural attributes above,
we encode them into $K=7$ part-aligned structural tokens:
$T_{\text{category}}$,
$T_{\text{semantic},i}$,
joint token (including $l_i$,
$J_i$),
$T_{\text{conn},i}$,
$\bs{c}_{\text{box}, i}$,
$\bs{s}_{\text{box}, i}$,
cavity token (including $c_{\text{cav}, i}$,
$\bs{s}_{\text{cav}, i}$).

Let $E_i^{\mathrm{str}}\in\mathbb{R}^{7\times d_s}$ denote
these structural tokens. Meanwhile, we employ a shared frozen DINOv2 encoder~\citep{dinov2} to extract visual features from the input image $I_0$ and the optional open-state reference $I_{\mathrm{open}}$ generated by GPT-image-2, which provides complementary cues for occluded internal structures, yielding $\mathbf{E}^{\mathrm{vis}}=\left[\mathcal{E}_{\mathrm{img}}(I_0);\mathcal{E}_{\mathrm{img}}(I_{\mathrm{open}})\right]$.

Generation proceeds through four steps with two complementary conditioning mechanisms:

\noindent\textit{(1) Multi-Part Latent Initialization.}
Given the predicted number of parts $N$, we maintain one group of geometry latent tokens for each part. At sampling timestep $t$, the latent representations of all parts are organized as
\begin{equation}
\mathbf{Z}_t =
[\mathbf{Z}_t^{1};
\mathbf{Z}_t^{2};
\ldots;
\mathbf{Z}_t^{N}],
\qquad
\mathbf{Z}_t^{i}\in\mathbb{R}^{M\times D},
\end{equation}
where each $\mathbf{Z}_t^i$ remains aligned with part $i$ throughout the denoising process. The latent tokens of each part are first projected into the DiT hidden space and combined with the timestep embedding:
\begin{equation}
\mathbf{H}_0^i
=
f_{\mathrm{in}}(\mathbf{Z}_t^i)
+
\mathbf{e}_t,
\end{equation}
where $\mathbf{e}_t$ denotes the timestep embedding.

\noindent\textit{(2) Per-Layer Structural Bias Injection (\emph{E}).}
At each DiT layer $l$, the seven part-aligned structural tokens
$\mathbf{E}_i^{\mathrm{str}}\in\mathbb{R}^{7\times d_s}$
are first aggregated into a part-level structural summary:
\begin{equation}
\bar{\mathbf{e}}_i^{\mathrm{str}}
=
\frac{1}{K}
\sum_{k=1}^{K}
\mathbf{E}_{i,k}^{\mathrm{str}},
\qquad K=7.
\end{equation}
The summary feature is then projected into the DiT hidden space and added to the current hidden geometry tokens:
\begin{equation}
\widetilde{\mathbf{H}}_l^i
=
\mathbf{H}_l^i
+
f_E\!\left(\bar{\mathbf{e}}_i^{\mathrm{str}}\right).
\end{equation}
This provides a persistent part-level structural bias at every DiT layer.

\noindent\textit{(3) Local, Global, and Image Attention.}
The structurally biased geometry tokens are subsequently processed by the original Multi-Part DiT backbone. Specifically, local self-attention first models interactions among geometry tokens within each part, global self-attention enables information exchange across different parts, and image cross-attention conditions the geometry tokens on the visual features
$\mathbf{E}^{\mathrm{vis}}$ extracted from $I_0$ and the optional $I_{\mathrm{open}}$.
We compactly denote these backbone operations as
\begin{equation}
\widehat{\mathbf{H}}_l^i
=
\mathcal{B}_l
\left(
\widetilde{\mathbf{H}}_l^i,
\mathbf{E}^{\mathrm{vis}}
\right),
\end{equation}
where $\mathcal{B}_l$ denotes the original $l$-th DiT block containing local self-attention, global self-attention, and image cross-attention.

\noindent\textit{(4) Structural Cross-Attention (\emph{C}).}
After the original backbone operations, we further introduce fine-grained structural conditioning by directly attending to the full structural token sequence. The updated geometry tokens
$\widehat{\mathbf{H}}_l^i$ serve as queries, while
$\mathbf{E}_i^{\mathrm{str}}$ provides the keys and values:
\begin{equation}
\mathbf{Q}_i^C
=
\widehat{\mathbf{H}}_l^i\mathbf{W}_Q^C,
\qquad
\mathbf{K}_i^C
=
\mathbf{E}_i^{\mathrm{str}}\mathbf{W}_K^C,
\qquad
\mathbf{V}_i^C
=
\mathbf{E}_i^{\mathrm{str}}\mathbf{W}_V^C.
\end{equation}
The structural attention weights and the resulting hidden representation are computed as
\begin{equation}
\mathbf{A}_i^C
=
\operatorname{softmax}
\left(
\frac{
\mathbf{Q}_i^C(\mathbf{K}_i^C)^{\top}
}{
\sqrt{d}
}
\right),
\qquad
\mathbf{H}_{l+1}^i
=
\widehat{\mathbf{H}}_l^i
+
\mathbf{A}_i^C\mathbf{V}_i^C,
\end{equation}
where $d$ is the attention-head dimension. In contrast to \emph{E}, which provides a shared part-level structural bias, \emph{C} allows individual geometry tokens to dynamically retrieve the most relevant semantic, spatial, and kinematic information from $\mathbf{E}_i^{\mathrm{str}}$.

These operations are repeatedly applied across DiT layers and sampling timesteps, progressively transforming the initial noisy part latents into structured part-level geometry representations.

\subsection{Training and Failure-Driven Data Expansion}
The structural encoder, E/C modules, and attention LoRA parameters are trained
with the rectified-flow velocity loss
\begin{equation}
\mathcal L_{\mathrm{RF}}=\mathbb E\bigl[\|V_\theta(Z_t,\Phi,t)-U^*\|_2^2\bigr],
\end{equation}
Here, $t$ denotes the sampled flow timestep, $\mathbf{Z}_t$ is the
interpolated latent at timestep $t$, ${\Phi}$ denotes the combined visual
and structural conditions, $V_{\theta}$ is the predicted velocity field, and
$\mathbf{U}^{*}$ is the target velocity defined by the Rectified Flow
interpolation path. The learning rate is
$10^{-4}$ and effective batch size is 8; timesteps follow a logit-normal
sample with $\mu=0$, $\sigma=1$. The initial approximately 1,100
programmatically modeled assets support 10,000 training steps. Targeted
modeling of failure categories adds approximately 1,300 assets, followed by
15,000 more steps. The reported total is approximately 2,400 assets and
25,000 steps, taking about 80 hours on four NVIDIA A100 GPUs.

\subsection{Binding Memory and Validation Records}
The binding agent retrieves category-level motion summaries and may expand
them into joint-specific axes, pivots, types, and ranges. This hierarchical
articulation memory supplies priors rather than proof of correctness.
Part identifiers and the predicted tree bind geometry to URDF links.
Isaac Sim motion renders are inspected for axis, direction, pivot, range,
and interference errors. Parameter repair and rollback precede fallback to
structure planning or geometry generation. Records retain the candidate
structure, geometry, motion render, critique, and adoption or rejection reason.
Quantitative comparisons are collected in Appendix~\ref{app:results-isart}.

\subsection{Asset Registry Fields}
\label{app:asset-registry}

The Asset Registry preserves both adoption and negative evidence.  A rejected
candidate, placeholder, or missing collision representation remains visible in
the audit even when another candidate is accepted.

\begin{longtable}{@{}p{0.29\linewidth}p{0.63\linewidth}@{}}
\caption{Conceptual asset-registry contract.}\label{tab:asset-registry}\\
\toprule
Group & Required audit meaning \\
\midrule
\endfirsthead
\toprule
Group & Required audit meaning \\
\midrule
\endhead
Request identity & Stable request, candidate, and instance identities that survive acquisition, placement, and export. \\
Semantic request & Description, object role, dimensions, attributes, style context, and articulation requirements. \\
Lifecycle & Planned, attempted, adopted, rejected, failed, and superseded states with terminal reason. \\
Provenance & Retrieval or generation source, query/prompt summary, backend/version, and cache status. \\
Geometry contract & Canonical orientation, scale, visual geometry, collision geometry, decomposition audit, and alignment evidence. \\
Material and physics & Material references, mass/friction properties, and simulator-readiness status. \\
Validation history & Warnings, hard failures, candidate comparisons, and evidence used for the adoption decision. \\
\bottomrule
\end{longtable}

%% file: sections/appendix_unravel.tex
\section{Unravel}
\label{app:unravel}

\subsection{Verified Localization and Deconstruction}
\label{app:unravel-boxes}
The planner proposes boxes for all targets in a global call. Boxes of the same
category with IoU above 0.85 are treated as duplicates. Each target receives
an independent local probe; IoU of at least 0.5 confirms the global box.
Otherwise, three independent local detections are sampled, and detections
agreeing with another sample at IoU of at least 0.7 are retained. Their
element-wise median replaces the drifting box. Failed probes or local
detections preserve the global estimate. A further inspection adds missed
movable objects to the protected set, preventing accidental deletion.

\begin{figure}[t]
\centering
\includegraphics[width=\linewidth]{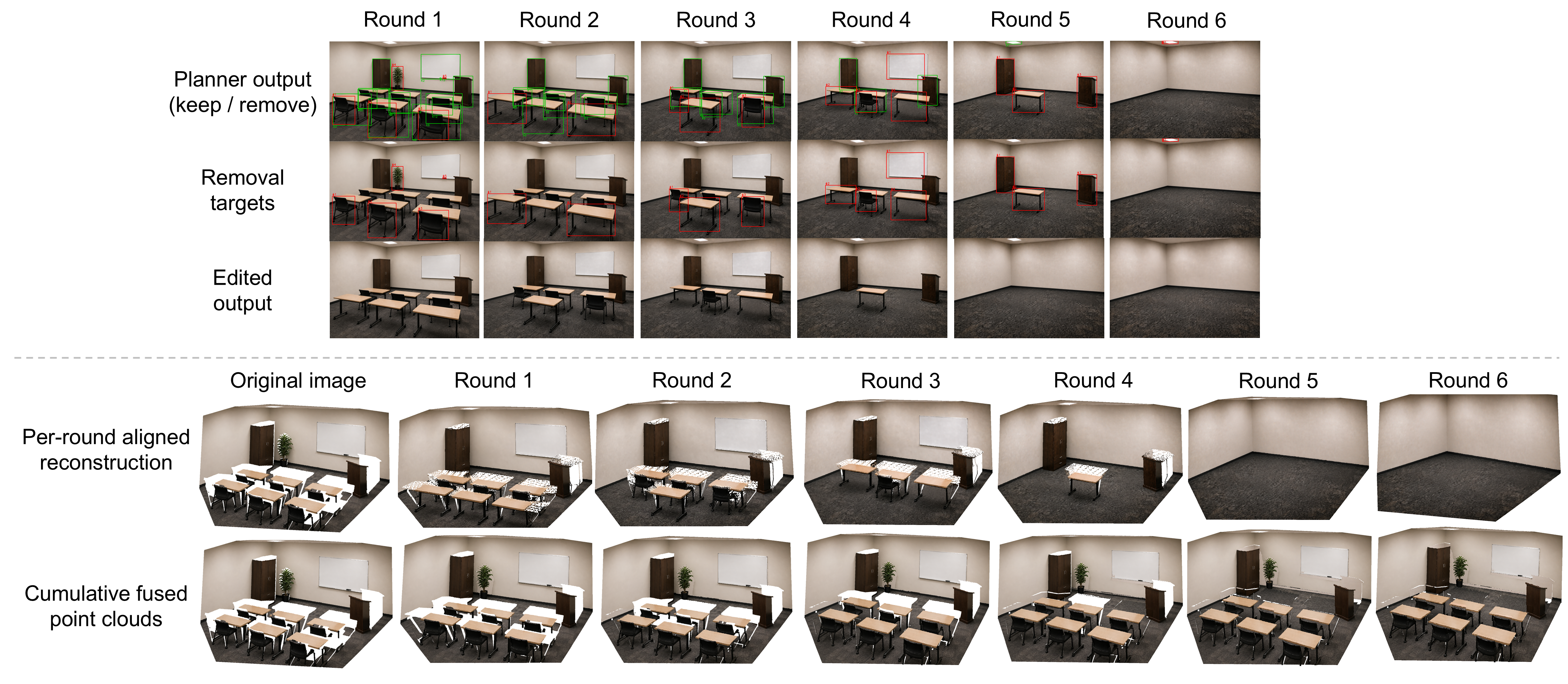}
\caption{Layer-wise deconstruction and point-cloud fusion in Unravel.
The upper block shows planner outputs, removal targets, and edited images
across six rounds. Red boxes mark removal targets; green boxes mark protected
objects. The lower block shows aligned per-round reconstructions and
cumulative fused point clouds.}
\label{fig:unravel-deconstruction-detail}
\end{figure}

\subsection{Point Fusion, Registration, and Settling}
\label{app:unravel-registration}
Point-map validity uses confidence threshold 0.5 and excludes unreliable
boundary pixels. Every round aligns to the first round through never-edited
anchors. Only newly exposed valid points enter fusion, retaining per-point
pixel and round provenance. Per-ray removal order assigns the fused points
to independent object clusters or the room shell.

Registration ranks transformations by bidirectional median Chamfer distance,
then refines trusted correspondences with truncated similarity ICP. A global
scale and bounded object-frame axis adjustments avoid interpreting occluded
extent as true object size. Under-observed axes may expand but cannot shrink
below observed span. Well-observed axes may shrink if fit does not worsen.
When all axes have insufficient observed extent, uniform scaling preserves
shape proportions.

Mounted objects remain fixed. Movable objects receive translations that
minimize total squared displacement subject to non-interpenetration, support,
and consistency with their attributed observation clusters. A supporting
surface lies beneath the object's local footprint. With bottom height $b_i$
and eligible support heights $H_i$, contact requires
\begin{equation}
0\le b_i-\max H_i\le\epsilon_s,\qquad\epsilon_s=0.03\,\mathrm m.
\end{equation}
The floor is always an eligible support. Observation-consistent translations
satisfy $d(a_i',C_i)\le\max(d(a_i,C_i),\epsilon)$, with tolerance
$\epsilon=0.3$ m horizontally and $0.15$ m vertically. These are tolerances
relative to observation fit, not a universal bound on every displacement.
Audit edits are followed by renewed feasibility checks.

\subsection{Support and Coordinate Records}
\label{app:support-coordinate}
Asset-local, camera, world, collision, and simulator frames retain their
transform relations and declared scale. Support uses measured local geometry
under each object footprint. Where a supporting parent exists, its instance
identity and supporting surface are recorded; semantic occlusion order alone
cannot establish contact. Mounted objects and articulation links preserve
their fixed or link-local transforms. Registration and settling records retain
observation fit, support gaps, penetrations, and accepted pose corrections.
Module results and blind scoring details are in
Appendix~\ref{app:results-unravel}.

%% file: sections/appendix_policyforge.tex
\section{PolicyForge}
\label{app:policyforge}

PolicyForge generates robot demonstrations from validated worlds and task
contracts. Source-specific robot, camera, action, and reset adapters preserve
the provenance required to admit trajectories into co-training. World
acceptance and episode success are evaluated separately.

\subsection{World-to-Policy Contract}
\label{app:world-policy-contract}

The policy interface is intentionally algorithm-neutral.  A policy module may
consume an accepted world only after the scene, task, simulator, and reset
contracts have been validated.  The following setup documents the evaluated WAM co-training configuration;
real-robot transfer is not measured by the reported experiments.

\begin{longtable}{@{}p{0.25\linewidth}p{0.67\linewidth}@{}}
\caption{Conceptual world-to-policy handoff.}\label{tab:world-policy-contract}\\
\toprule
Component & Required contents \\
\midrule
\endfirsthead
\toprule
Component & Required contents \\
\midrule
\endhead
World & Accepted scene state, strict validation summary, simulator package, coordinate contract, and provenance. \\
Task & Task family, goal, ordered subgoals, target entities, affordances, joint targets, and success predicate. \\
Robot adapter & Embodiment, sensors, observation/action spaces, control horizon, and reset assumptions. \\
Randomization & Scene, asset, reset, and rollout seeds plus declared domain-randomization policy. \\
Rollout feedback & Terminal status, first blocker, owner, evidence, and recovery decision returned to the Harness. \\
\bottomrule
\end{longtable}

\subsection{Rollout Metadata}
\label{app:rollout-metadata}

Each rollout is joinable to one world, task, policy, robot, simulator, seed,
and checkpoint.  The minimum conceptual record is summarized below; exact
serialization is left to the policy owner.

\begin{longtable}{@{}p{0.25\linewidth}p{0.67\linewidth}@{}}
\caption{Conceptual rollout record.}\label{tab:rollout-metadata}\\
\toprule
Group & Required meaning \\
\midrule
\endfirsthead
\toprule
Group & Required meaning \\
\midrule
\endhead
Identity & World, task, policy, robot, simulator, cohort split, and checkpoint identifiers. \\
Randomness and budget & Scene/reset/rollout seeds, horizon, environment steps, and declared resource budget. \\
Outcome & Success predicate, terminal status, return/length/latency summaries, and infrastructure validity. \\
Failure attribution & First blocker, causal owner, affected entity or surface, secondary failures, and recovery evidence. \\
\bottomrule
\end{longtable}

\subsection{Policy-Learning Setup}
\label{sec:appendix-setup}

\subsubsection{WAM Architecture}
\label{sec:appendix-wam-arch}

WAM adopts a dual-stream Video DiT + Action DiT architecture,
initialized from the Wan2.2-TI2V-5B pretrained video generation model.

\begin{table}[ht]
\centering
\caption{WAM architecture configuration.}
\label{tab:wam-arch}
\small
\begin{tabularx}{\linewidth}{l|X}
\toprule
\textbf{Component} & \textbf{Configuration} \\
\midrule
Video DiT        & 30 layers, $d_\text{hidden}=3072$, $d_\text{ffn}=14336$, 24 heads ($d_\text{head}=128$) \\
Action DiT       & 30 layers, $d_\text{hidden}=1024$, $d_\text{ffn}=4096$, 24 heads ($d_\text{head}=128$) \\
Text Encoder     & UMT5-XXL (frozen, precomputed embeddings, context\_len\,=\,128) \\
Input Cameras    & 2 cameras (front + wrist), $224 \times 224$ each, concat $\rightarrow$ $224 \times 448$ \\
Video Context    & 33 frames, action\_video\_freq\_ratio\,=\,4 \\
Action Space     & 7-DoF: 6 delta ($xyz$ + $rpy$) + 1 absolute gripper ($0$=close, $1$=open) \\
Proprioception   & 8-dim state vector \\
Normalization    & min/max per-dimension \\
Diffusion        & Flow matching, train\_shift\,=\,5.0, infer\_shift\,=\,5.0, 1000 timesteps \\
Loss             & $\lambda_\text{action} = 1.0$ \\
\bottomrule
\end{tabularx}
\end{table}

\subsubsection{Training Hyperparameters}
\label{sec:appendix-training-hp}

LIBERO-Pro experiments are trained on a single node with 8 GPUs;
LIBERO-Plus experiments scale to 4 nodes (32 GPUs), with the learning rate
adjusted according to the square-root scaling rule.

\begin{table}[ht]
\centering
\caption{Training hyperparameters. The LIBERO-Plus configuration uses
$\sqrt{6} \times 10^{-4} \approx 2.449 \times 10^{-4}$ to compensate
for the $6 \times$ larger global batch size.}
\label{tab:training-hp}
\small
\begin{tabular}{l|cc}
\toprule
\textbf{Hyperparameter} & \textbf{LIBERO-Pro (1-node)} & \textbf{LIBERO-Plus (4-node)} \\
\midrule
GPUs                      & 8 ($1 \times 8$)       & 32 ($4 \times 8$) \\
Batch size (per GPU)      & 16                      & 24 \\
Global batch size         & 128                     & 768 \\
Learning rate             & $1 \times 10^{-4}$      & $2.449 \times 10^{-4}$ ($= \sqrt{6} \times 10^{-4}$) \\
LR schedule               & Cosine                  & Cosine \\
Epochs                    & 10                      & 10 \\
Max steps                 & ---                     & 15K \\
Optimizer                 & AdamW (DeepSpeed ZeRO-1) & AdamW (DeepSpeed ZeRO-1) \\
Weight decay              & $1 \times 10^{-2}$      & $1 \times 10^{-2}$ \\
Gradient accumulation     & 1                       & 1 \\
Precision                 & bf16 mixed              & bf16 mixed \\
Checkpoint save interval  & 2{,}000 steps           & 1{,}000 steps \\
\bottomrule
\end{tabular}
\end{table}

\subsubsection{Co-training Data Configuration}
\label{sec:appendix-data-config}

\paragraph{Data sources.}
LIBERO~\citep{liu2023libero} is an open-source benchmark dataset based on MuJoCo.
Our synthetic demonstrations are generated from Awomo-SimDataEngine scenes using
PolicyForge in Isaac Sim and organized into three data families:
\textsc{SimData-Easy}, with simple scenes and tasks;
\textsc{SimData-Lite}, with simple scenes and more challenging tasks;
and \textsc{SimData-Hard}, with richer scenes and challenging tasks.
\textsc{SimData-Ultra} denotes a co-training configuration combining
multiple data sources with robot initial-state and language augmentation.

Co-training mixes benchmark demonstrations and additional data using sampling
weights, with a reported 1:1 effective-frame ratio. SimData-Ultra combines
SimData-Hard, additional in-house demonstrations from complex scenes, and
LIBERO-Plus data with robot initial-state and language augmentation.
Exact source weights and augmentation settings must be retained
in the run manifest; we do not enumerate them fully in this report.

\begin{table}[ht]
\centering
\caption{Co-training data mixing for LIBERO-Plus experiments.
LIBERO-Plus provides the benchmark demonstrations.}
\label{tab:data-mixing-libero-plus}
\small
\begin{tabularx}{\linewidth}{l|Xlc}
\toprule
\textbf{Config} & \textbf{Data Description} & \textbf{Sampling Weight} & \textbf{Frame Ratio} \\
\midrule
Baseline         & LIBERO-Plus only & --- & --- \\
+ SimData-Hard  & Richer scenes with challenging tasks & 0.308 & 1\,:\,1 \\
+ SimData-Ultra & Multi-source mixture with robot initial-state and language augmentation & varied & 1\,: \,1 \\
\bottomrule
\end{tabularx}
\end{table}

\subsubsection{Evaluation Protocol}
\label{sec:appendix-eval}

\begin{table}[ht]
\centering
\caption{Evaluation protocol summary for all benchmarks.}
\label{tab:eval-protocol}
\small
\begin{tabularx}{\linewidth}{l|ccX}
\toprule
\textbf{Benchmark} & \textbf{Simulator} & \textbf{Scale} & \textbf{Details} \\
\midrule
LIBERO-Plus  & MuJoCo & 1{,}003 episodes
  & 4 suites (Obj/Spa/Goal/L-10), 8 GPU / 16 workers \\
LIBERO-Pro   & MuJoCo & 1{,}000 episodes
  & 200 configs $\times$ 5 trials, 4 suites $\times$ 5 dimensions \\
\bottomrule
\end{tabularx}
\end{table}

The tables list available evaluated checkpoints. The main LIBERO-Plus
comparison selects each configuration's highest-overall-success evaluated
checkpoint up to and including 13K steps; all suite scores use that same
selected checkpoint. Baseline runs finish at 15K and the recorded
Ultra runs at 14,580; unlisted final results are not imputed.
WAM evaluations on LIBERO-Plus use 1,003 episodes.
On LIBERO-Pro, Ultra peaks at 48.7\% at 13K steps and finishes at
48.0\% at 14,580 steps. The final per-dimension and per-suite results
in Tables~\ref{tab:libero-pro-dim} and~\ref{tab:libero-pro-suite}
use Ultra's 14,580-step checkpoint rather than its peak checkpoint.


%% file: sections/appendix_results.tex
\section{Experimental Results}
\label{app:experimental-results}

The module evaluations measure asset geometry, articulation binding, and
single-view reconstruction. Policy evaluations measure downstream utility of
synthetic demonstrations. These are separate protocols and are not pooled
into a single system score. The tables preserve evaluation conditions and
checkpoint identities; unreported values are not inferred.

\subsection{ISArt Results}
\label{app:results-isart}
The benchmark contains 60 objects: 30 from World-SimReady-Home
\citep{isart_yootta} and 30 from GAPartNet\_PhyScene
\citep{isart_gapartnet}, spanning 12 categories. Geometry methods use the
same input images and are compared with corresponding ground-truth part
meshes. ISArt adopts object-level normalization, where the bounding-box center of the
entire object is translated to the origin and all parts are uniformly scaled
using a shared scale factor. We evaluate geometry using F-Score at thresholds
0.05 and 0.1 (F@0.05 and F@0.1) and Chamfer Distance (CD). For each semantically
corresponding component pair, we sample 10,000 surface points per side and
compute the geometry metrics. Scores are averaged over three sampling repeats,
component pairs, and objects.The results are reported in in Table~\ref{tab:isart-part-geometry}.

\begin{table}[ht]
\centering\small
\caption{ISArt part geometry on the 60-object benchmark.}
\label{tab:isart-part-geometry}
\begin{tabularx}{\linewidth}{Xccc}
\toprule
Method & CD $\downarrow$ & F@0.05 $\uparrow$ & F@0.1 $\uparrow$\\
\midrule
PartCrafter & 0.421 & 0.229 & 0.395\\
OmniPart & 0.305 & 0.306 & 0.540\\
SPARK & 0.278 & 0.314 & 0.505\\
TRELLIS.2 + P3-SAM + X-Part & 0.268 & 0.431 & 0.627\\
ISArt & \textbf{0.163} & \textbf{0.604} & \textbf{0.789}\\
\bottomrule
\end{tabularx}
\end{table}

Binding is evaluated with ground-truth part geometry (using the same dataset as above) to isolate kinematic
estimation from reconstruction.Following the ISArt binding evaluation protocol, AxisErr measures joint-axis
deviation, PivotErr measures pivot-position error, and TypeErr denotes the
percentage of incorrectly predicted joint types. The reported success rate concerns binding with supplied geometry.The results are reported in in Table~\ref{tab:isart-binding}.

\begin{table}[ht]
\centering\small
\caption{ISArt URDF binding with fixed ground-truth geometry.}
\label{tab:isart-binding}
\begin{tabularx}{\linewidth}{Xcccc}
\toprule
Method & AxisErr $\downarrow$ & PivotErr $\downarrow$ & TypeErr $\downarrow$ & Success $\uparrow$\\
\midrule
Articulate AnyMesh & 0.72 & \textbf{0.02} & 46.67\% & 80\%\\
Articulate-Anything & 0.12 & 0.17 & 21.11\% & 90\%\\
SPARK & 0.38 & 0.19 & 23.61\% & \textbf{100\%}\\
ISArt & \textbf{0.07} & 0.05 & \textbf{13.33\%} & \textbf{100\%}\\
\bottomrule
\end{tabularx}
\end{table}

\begin{figure}[t]
  \centering
  \includegraphics[width=\linewidth]{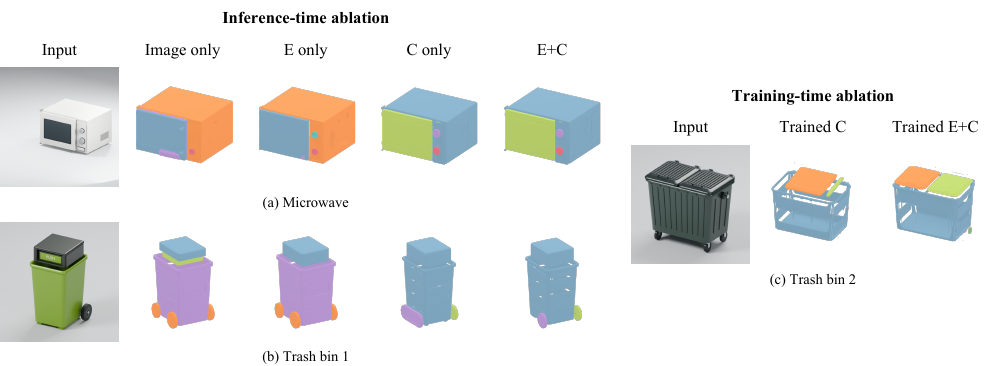}
  \caption{ISArt inference and training ablations for structural conditioning E and C. The examples support complementary effects; no aggregate ablation metric is reported.}
  \label{fig:isart-conditioning-ablation}
\end{figure}

\paragraph{Conditioning ablations.}
We qualitatively examine the effects of the structural conditioning branches $E$ and $C$. 
At inference time, we independently enable $E$, $C$, or $E{+}C$ from the same trained checkpoint to visualize their effects on generated part geometry. 
We further conduct a controlled training-time comparison by training $C$-only and $E{+}C$ variants from the same backbone for 1,000 steps on 100 trash-bin samples. 
As shown in Figure~\ref{fig:isart-conditioning-ablation}, $C$ mainly improves global part layout and structural consistency, while $E$ provides complementary local-detail refinement.

\FloatBarrier
\subsection{Unravel Results}
\label{app:results-unravel}
The controlled deconstruction comparison uses five scenes with 160 objects.
Direct reconstruction and Unravel share SAM 3 segmentation, TRELLIS.2
asset generation, and MoGe3 point estimation; the direct variant only uses
the input image. Unravel fuses four to eight deconstruction rounds.
The pipeline components are used off the shelf without fine-tuning.

\begin{table}[ht]
\centering\small
\caption{Unravel deconstruction comparison. Mask area and under-coverage
rates (UCR, coverage below 70\%) apply to objects removed after the first
round, using Unravel masks and clusters as references. Fidelity scores
average all 160 objects; point counts average scenes.}
\label{tab:unravel-ablation}
\setlength{\tabcolsep}{4pt}
\begin{tabularx}{\linewidth}{Xcccccc}
\toprule
& \multicolumn{2}{c}{Masks} & \multicolumn{2}{c}{Assets (1--10)} & \multicolumn{2}{c}{Geometry}\\
\cmidrule(lr){2-3}\cmidrule(lr){4-5}\cmidrule(lr){6-7}
Method & Area ratio & UCR (\%) & VLM & Human & Points (M) & UCR (\%)\\
\midrule
Direct & 0.772 & 11.3 & 4.7 & 4.9 & 1.53 & 10.1\\
Unravel & 1.000 & 0.0 & 8.2 & 8.6 & 2.81 & 0.0\\
\bottomrule
\end{tabularx}
\end{table}

\paragraph{Coverage interpretation.}
Objects removed after the first round account for approximately one third
of the test set. Their direct mask area averages 77.2\% of the Unravel
reference, and direct point coverage averages 76.9\%. The zero UCR of
Unravel follows from using its own masks and clusters as the reference.
Additional points include geometry predicted from inpainted surfaces.
Consequently these statistics measure relative coverage, not accuracy against
independent ground-truth geometry.

\paragraph{Blind asset-fidelity scoring.}
Each object supplies two meshes, rendered from identical four views separated
by $90^\circ$. Paired renders are randomly ordered without revealing the
pipeline. Both are judged against the crop from the object's best-observed
round. Human raters and GPT-5.5 score independently on the same rubric:
1--3 for wrong shape or missing essential parts; 4--6 for recognizable but
deformed or incomplete objects; 7--8 for correct overall shape with minor
artifacts; and 9--10 for faithful shape and structure. Scores are means over
all 160 objects. We do not report confidence intervals or a
rater-agreement statistic, and the reference may include generated completion.

\begin{figure}[t]
\centering
\includegraphics[width=\linewidth]{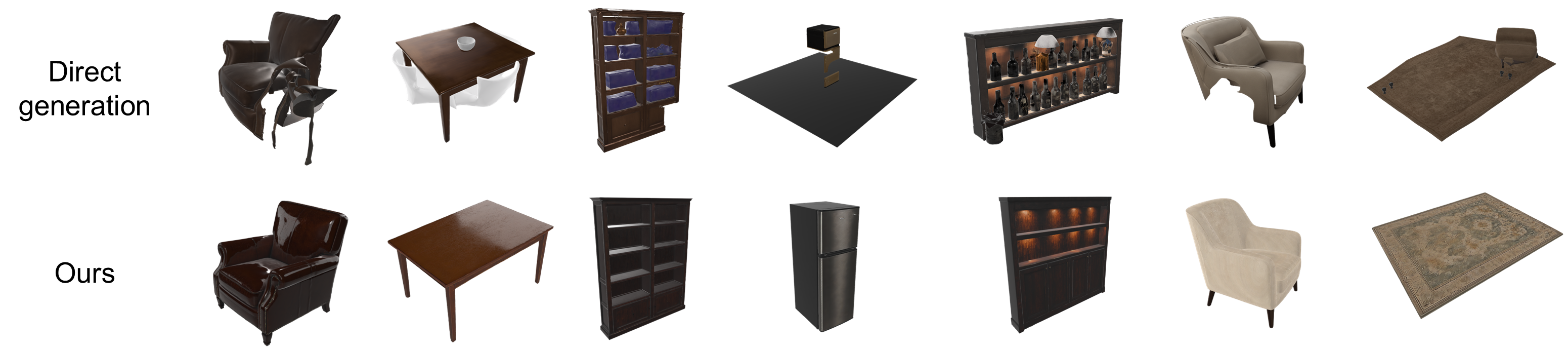}
\caption{Asset examples generated directly from the input masks (top) and
from the best-observed deconstruction round (bottom).}
\label{fig:unravel-asset-ablation}
\end{figure}

\begin{figure}[p]
\centering
\includegraphics[width=\linewidth,height=0.77\textheight,keepaspectratio]{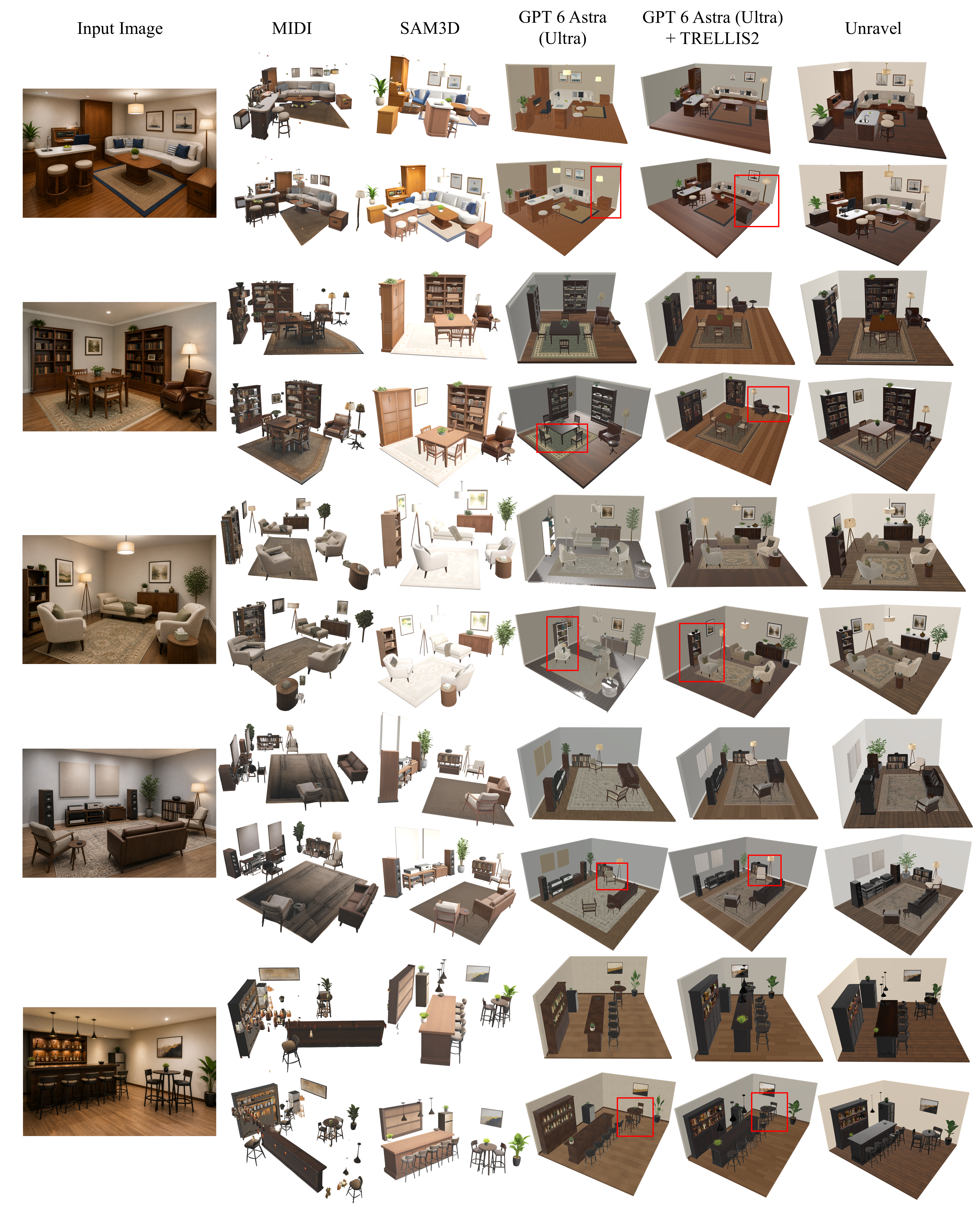}
\caption{Single-image scene comparisons. The shared-asset agentic variant
receives Unravel's meshes and masks, isolating pose and scale estimation.}
\label{fig:unravel-baselines}
\end{figure}

\begin{figure}[p]
\centering
\includegraphics[width=\linewidth,height=0.77\textheight,keepaspectratio]{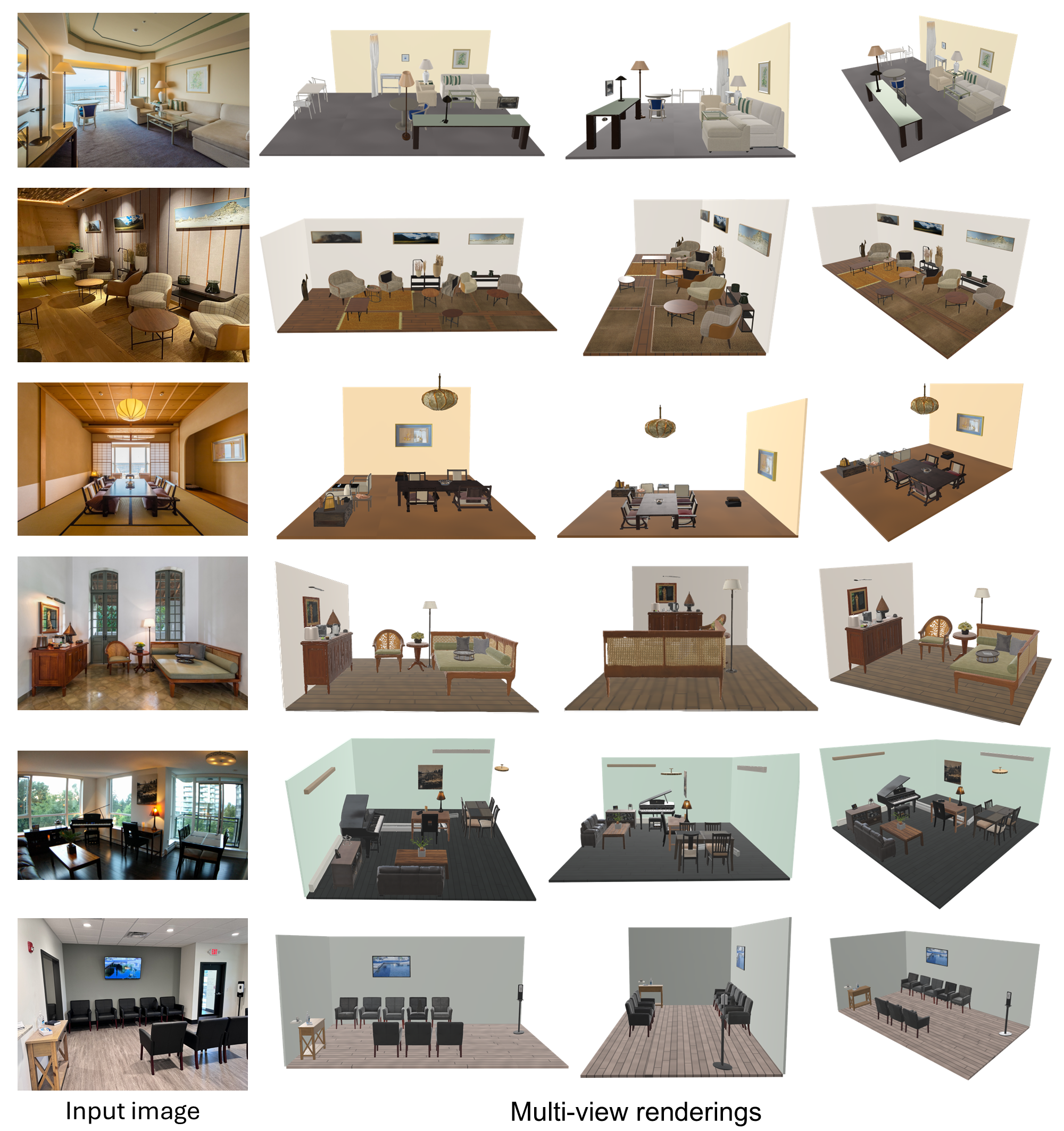}
\caption{Unravel reconstructions from six real indoor photographs. Each row
shows the input and three reconstructed views.}
\label{fig:unravel-real-results}
\end{figure}

\paragraph{Placement and scene examples.}
The shared-asset agentic baseline receives the same generated meshes and
2D masks as Unravel, estimating only scale and pose. This comparison isolates
placement from mesh-generation quality (Figure~\ref{fig:unravel-baselines}).
We also present Unravel reconstructions from six real indoor photographs
(Figure~\ref{fig:unravel-real-results}); these examples are qualitative and
are not included in the five-scene quantitative aggregate.

\begin{figure}[p]
\centering
\includegraphics[width=\linewidth,height=0.77\textheight,keepaspectratio]{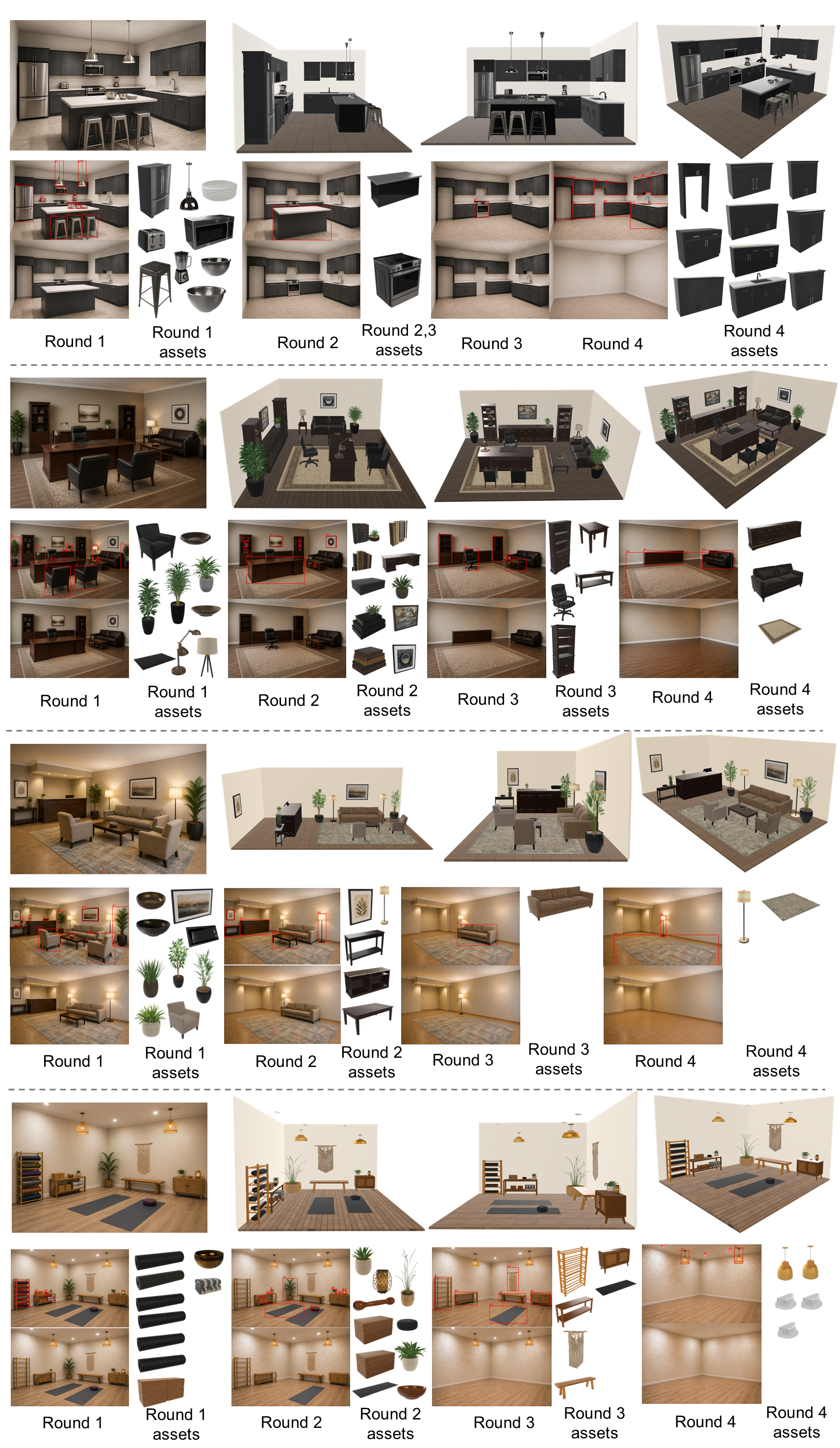}
\caption{Additional Unravel scenes with input images, reconstructed views,
per-round edits, and extracted assets. These traces are qualitative examples.}
\label{fig:unravel-additional-scenes}
\end{figure}

\clearpage
\subsection{Policy Co-training Results}

\label{sec:appendix-data}

The tables below retain the recorded checkpoint results, including fixed-13K
and final-checkpoint summaries. The main-text LIBERO-Plus comparison instead
selects each configuration's highest-overall-success evaluated checkpoint
up to and including 13K steps (Table~\ref{tab:exp-plus-main}); its suite
scores all come from that selected checkpoint.


\begin{table}[ht]
\centering
\caption{LIBERO-Plus overall success rate (\%) across training steps.
Co-training models mix the corresponding additional data family at
equal effective-frame ratio; the baseline uses benchmark data only. Evaluation: 1{,}003 episodes,
4 generalization suites. The final column uses 15K for the baseline and 14,580 for Ultra;
final results for the other sources are not listed. Best per-step result in \textbf{bold}.}
\label{tab:libero-plus-overall}
\begin{tabular}{l|cccccccc}
\toprule
\textbf{Co-training Data} & \textbf{1K} & \textbf{3K} & \textbf{5K} & \textbf{7K} & \textbf{9K} & \textbf{11K} & \textbf{13K} & \textbf{Final} \\
\midrule
Baseline (no synthetic)    & 35.49 & 62.61 & 64.91 & 71.39 & 77.17 & 76.57 & 75.97 & 74.28 \\
+ SimData-Easy            & 48.65 & 59.52 & 69.89 & 74.48 & 77.17 & 76.27 & 78.76 & --- \\
+ SimData-Lite            & 40.08 & 59.32 & 68.20 & 73.28 & 75.97 & 77.77 & 78.27 & --- \\
+ SimData-Hard            & 36.89 & 63.11 & 74.38 & 72.78 & 79.46 & 79.36 & 81.06 & --- \\
+ SimData-Ultra & \textbf{65.90} & \textbf{81.95} & \textbf{84.75} & \textbf{88.14} & \textbf{86.94} & \textbf{87.64} & \textbf{89.43} & \textbf{89.63} \\
\bottomrule
\end{tabular}
\end{table}


\begin{table}[ht]
\centering
\caption{LIBERO-Plus per-suite success rate (\%) at training step 13{,}000.
All configurations evaluated at the same checkpoint step for fair comparison.
Best per-suite result in \textbf{bold}.}
\label{tab:libero-plus-suite-13k}
\begin{tabular}{l|ccccc}
\toprule
\textbf{Co-training Data}
  & \textbf{Overall}
  & \textbf{Object}
  & \textbf{Spatial}
  & \textbf{Goal}
  & \textbf{LIBERO-10} \\
\midrule
Baseline (no synthetic)   & 75.97 & 84.13 & 81.25 & 54.83 & 84.52 \\
+ SimData-Easy           & 78.76 & 86.51 & 85.83 & 58.30 & 85.32 \\
+ SimData-Lite           & 78.27 & 85.71 & 83.33 & 61.78 & 82.94 \\
+ SimData-Hard           & 81.06 & 87.70 & 91.25 & 64.09 & 82.14 \\
+ SimData-Ultra & \textbf{89.43} & \textbf{92.86} & \textbf{93.33} & \textbf{84.17} & \textbf{87.70} \\
\bottomrule
\end{tabular}
\end{table}


\begin{table}[ht]
\centering
\caption{LIBERO-Plus per-suite success rate (\%) for all configurations
across training steps. Suites: Object generalization (Obj),
Spatial generalization (Spa), Goal generalization (Goal),
Long-horizon sequential tasks (L-10).}
\label{tab:libero-plus-full}
\small
\begin{tabular}{ll|ccccc}
\toprule
\textbf{Co-training Data} & \textbf{Step}
  & \textbf{Overall} & \textbf{Obj} & \textbf{Spa} & \textbf{Goal} & \textbf{L-10} \\
\midrule
\multirow{8}{*}{Baseline}
  & 1K  & 35.49 & 55.16 & 41.67 & 24.32 & 21.43 \\
  & 3K  & 62.61 & 84.52 & 71.25 & 37.07 & 58.73 \\
  & 5K  & 64.91 & 79.37 & 72.08 & 39.77 & 69.44 \\
  & 7K  & 71.39 & 83.73 & 81.25 & 51.74 & 69.84 \\
  & 9K  & 77.17 & 84.52 & 87.08 & 52.51 & 85.71 \\
  & 11K & 76.57 & 80.16 & 85.42 & 61.39 & 80.16 \\
  & 13K & 75.97 & 84.13 & 81.25 & 54.83 & 84.52 \\
  & 15K & 74.28 & 83.73 & 78.33 & 54.44 & 81.35 \\
\midrule
\multirow{7}{*}{+ SimData-Easy}
  & 1K  & 48.65 & 74.60 & 61.25 & 29.73 & 30.16 \\
  & 3K  & 59.52 & 74.60 & 69.58 & 47.10 & 47.62 \\
  & 5K  & 69.89 & 86.51 & 81.25 & 45.95 & 67.06 \\
  & 7K  & 74.48 & 86.51 & 81.67 & 50.19 & 80.56 \\
  & 9K  & 77.17 & 87.30 & 85.42 & 51.35 & 85.71 \\
  & 11K & 76.27 & 83.73 & 83.75 & 55.21 & 83.33 \\
  & 13K & 78.76 & 86.51 & 85.83 & 58.30 & 85.32 \\
\midrule
\multirow{7}{*}{+ SimData-Lite}
  & 1K  & 40.08 & 61.51 & 54.58 & 27.80 & 17.46 \\
  & 3K  & 59.32 & 72.22 & 71.67 & 45.17 & 49.21 \\
  & 5K  & 68.20 & 84.92 & 73.75 & 44.79 & 70.24 \\
  & 7K  & 73.28 & 88.10 & 78.75 & 54.44 & 72.62 \\
  & 9K  & 75.97 & 87.70 & 82.92 & 55.60 & 78.57 \\
  & 11K & 77.77 & 87.30 & 81.25 & 59.46 & 83.73 \\
  & 13K & 78.27 & 85.71 & 83.33 & 61.78 & 82.94 \\
\midrule
\multirow{7}{*}{+ SimData-Hard}
  & 1K  & 36.89 & 59.92 & 54.17 & 22.01 & 12.70 \\
  & 3K  & 63.11 & 84.92 & 75.42 & 43.24 & 50.00 \\
  & 5K  & 74.38 & 86.90 & 85.83 & 58.30 & 67.46 \\
  & 7K  & 72.78 & 87.30 & 82.92 & 51.35 & 70.63 \\
  & 9K  & 79.46 & 87.30 & 86.67 & 62.93 & 81.75 \\
  & 11K & 79.36 & 86.90 & 87.50 & 60.62 & 83.33 \\
  & 13K & 81.06 & 87.70 & 91.25 & 64.09 & 82.14 \\
\midrule
\multirow{8}{*}{+ SimData-Ultra}
 & 1K & 65.90 & 92.46 & 76.25 & 57.14 & 38.49 \\
 & 3K & 81.95 & 90.08 & 87.92 & 78.38 & 71.83 \\
 & 5K & 84.75 & 90.87 & 90.83 & 77.61 & 80.16 \\
 & 7K & 88.14 & 92.06 & 96.25 & 83.01 & 81.75 \\
 & 9K & 86.94 & 91.67 & 93.75 & 81.08 & 81.75 \\
 & 11K & 87.64 & 92.06 & 91.67 & 78.38 & 88.89 \\
 & 13K & 89.43 & 92.86 & 93.33 & 84.17 & 87.70 \\
 & 14580 & 89.63 & 92.46 & 92.50 & 84.56 & 89.29 \\
\bottomrule
\end{tabular}
\end{table}


\begin{table}[ht]
\centering
\caption{Improvement ($\Delta$\,pp) over Baseline at step 13{,}000
on LIBERO-Plus. Positive values indicate gains from co-training
with the corresponding synthetic data.}
\label{tab:libero-plus-delta-13k}
\begin{tabular}{l|ccccc}
\toprule
\textbf{Co-training Data}
  & \textbf{$\Delta$Overall}
  & \textbf{$\Delta$Object}
  & \textbf{$\Delta$Spatial}
  & \textbf{$\Delta$Goal}
  & \textbf{$\Delta$LIBERO-10} \\
\midrule
+ SimData-Easy  & +2.79 & +2.38 & +4.58 & +3.47 & +0.80 \\
+ SimData-Lite  & +2.30 & +1.58 & +2.08 & +6.95 & $-$1.58 \\
+ SimData-Hard  & +5.09 & +3.57 & +10.00 & +9.26 & $-$2.38 \\
+ SimData-Ultra & \textbf{+13.46} & \textbf{+8.73} & \textbf{+12.08} & \textbf{+29.34} & \textbf{+3.18} \\
\bottomrule
\end{tabular}
\end{table}


\begin{table}[ht]
\centering
\caption{LIBERO-Pro overall success rate (\%) across training steps.
Evaluation: 200 task configurations $\times$ 5 trials = 1{,}000 episodes
per checkpoint, spanning 4 base suites $\times$ 5 generalization dimensions.
Baseline final step is 15{,}000; all co-training configs final step is
14{,}580. Best per-step result in \textbf{bold}.}
\label{tab:libero-pro-overall}
\begin{tabular}{l|ccccc}
\toprule
\textbf{Co-training Data} & \textbf{1K} & \textbf{5K} & \textbf{9K} & \textbf{13K} & \textbf{Final} \\
\midrule
Baseline (no synthetic)  & 13.8 & 40.6 & 45.3 & 45.9 & 46.7 \\
+ SimData-Easy          & \textbf{15.8} & 42.2 & \textbf{47.7} & 47.6 & 47.9 \\
+ SimData-Lite          & 14.6 & 40.8 & 46.6 & 46.8 & 46.9 \\
+ SimData-Hard          & 13.1 & \textbf{43.3} & 45.7 & 46.9 & 47.1 \\
+ SimData-Ultra & --- & --- & 45.9 & \textbf{48.7} & \textbf{48.0} \\
\bottomrule
\end{tabular}
\end{table}


\begin{table}[ht]
\centering
\caption{LIBERO-Pro per generalization dimension success rate (\%) at the
final training step. Five dimensions: Object (novel objects), Swap
(position-swap), Language (semantic-language), Task (novel tasks),
Environment (novel environments). 200 episodes per dimension per config.
Best per-dimension result in \textbf{bold}.}
\label{tab:libero-pro-dim}
\begin{tabular}{l|ccccc}
\toprule
\textbf{Co-training Data}
  & \textbf{Object}
  & \textbf{Swap}
  & \textbf{Language}
  & \textbf{Task}
  & \textbf{Environ.} \\
\midrule
Baseline             & 79.50 & 3.00 & 93.50 & 18.00 & 39.50 \\
+ SimData-Easy      & 77.00 & \textbf{4.00} & 93.50 & \textbf{21.00} & 44.00 \\
+ SimData-Lite      & 77.00 & 3.00 & 90.50 & 18.50 & \textbf{45.50} \\
+ SimData-Hard      & \textbf{80.50} & \textbf{4.00} & 96.00 & 18.50 & 36.50 \\
+ SimData-Ultra & \textbf{80.50} & 1.00 & \textbf{97.00} & 16.50 & 45.00 \\
\bottomrule
\end{tabular}
\end{table}


\begin{table}[ht]
\centering
\caption{LIBERO-Pro per base suite success rate (\%) at the final
training step. Each suite contains 250 episodes (50 task configs
$\times$ 5 trials). Best per-suite result in \textbf{bold}.}
\label{tab:libero-pro-suite}
\begin{tabular}{l|ccccc}
\toprule
\textbf{Co-training Data}
  & \textbf{Overall}
  & \textbf{Object}
  & \textbf{Spatial}
  & \textbf{Goal}
  & \textbf{LIBERO-10} \\
\midrule
Baseline             & 46.70 & 47.60 & 55.20 & 44.00 & 40.00 \\
+ SimData-Easy      & 47.90 & 46.80 & 57.60 & 44.40 & \textbf{42.80} \\
+ SimData-Lite      & 46.90 & \textbf{48.80} & 53.60 & 46.00 & 39.20 \\
+ SimData-Hard      & 47.10 & \textbf{48.80} & 57.60 & 45.60 & 36.40 \\
+ SimData-Ultra & \textbf{48.00} & 45.60 & \textbf{60.40}
 & \textbf{48.80} & 37.20 \\
\bottomrule
\end{tabular}
\end{table}


\begin{table}[ht]
\centering
\caption{LIBERO-Pro per base suite success rate (\%) for all configurations
across training steps. Each suite contains 250 episodes per checkpoint.}
\label{tab:libero-pro-full}
\small
\begin{tabular}{ll|ccccc}
\toprule
\textbf{Co-training Data} & \textbf{Step}
  & \textbf{Overall} & \textbf{Obj} & \textbf{Spa} & \textbf{Goal} & \textbf{L-10} \\
\midrule
\multirow{5}{*}{Baseline}
  & 1K    & 13.80 & 16.80 & 20.80 & 12.00 &  5.60 \\
  & 5K    & 40.60 & 46.00 & 49.60 & 37.20 & 29.60 \\
  & 9K    & 45.30 & 47.60 & 53.20 & 40.80 & 39.60 \\
  & 13K   & 45.90 & 48.40 & 53.60 & 44.80 & 36.80 \\
  & 15K   & 46.70 & 47.60 & 55.20 & 44.00 & 40.00 \\
\midrule
\multirow{5}{*}{+ SimData-Easy}
  & 1K    & 15.80 & 24.00 & 25.20 & 12.80 &  1.20 \\
  & 5K    & 42.20 & 46.00 & 53.60 & 35.20 & 34.00 \\
  & 9K    & 47.70 & 50.00 & 58.00 & 40.80 & 42.00 \\
  & 13K   & 47.60 & 47.20 & 56.80 & 44.40 & 42.00 \\
  & 14580 & 47.90 & 46.80 & 57.60 & 44.40 & 42.80 \\
\midrule
\multirow{5}{*}{+ SimData-Lite}
  & 1K    & 14.60 & 23.20 & 22.80 & 10.80 &  1.60 \\
  & 5K    & 40.80 & 46.40 & 48.00 & 40.00 & 28.80 \\
  & 9K    & 46.60 & 50.00 & 53.20 & 45.60 & 37.60 \\
  & 13K   & 46.80 & 48.00 & 54.80 & 43.60 & 40.80 \\
  & 14580 & 46.90 & 48.80 & 53.60 & 46.00 & 39.20 \\
\midrule
\multirow{5}{*}{+ SimData-Hard}
  & 1K    & 13.10 & 16.80 & 21.20 & 12.00 &  2.40 \\
  & 5K    & 43.30 & 47.20 & 56.80 & 40.80 & 28.40 \\
  & 9K    & 45.70 & 46.00 & 55.60 & 46.40 & 34.80 \\
  & 13K   & 46.90 & 48.00 & 56.40 & 44.40 & 38.80 \\
  & 14580 & 47.10 & 48.80 & 57.60 & 45.60 & 36.40 \\
\midrule
\multirow{3}{*}{+ SimData-Ultra}
 & 9K & 45.90 & 45.20 & 57.60 & 47.20 & 33.60 \\
 & 13K & 48.70 & 46.00 & 60.80 & 48.80 & 39.20 \\
 & 14580 & 48.00 & 45.60 & 60.40 & 48.80 & 37.20 \\
\bottomrule
\end{tabular}
\end{table}

\begin{table}[htbp]
\centering
\small
\caption{Best listed WAM checkpoint per configuration on LIBERO-Plus and LIBERO-Pro.
Ultra's LIBERO-Pro peak is at 13K, not its final checkpoint.}
\label{tab:summary-all}
\begin{tabular}{lcc}
\toprule
Co-training data & LIBERO-Plus (WAM) & LIBERO-Pro (WAM) \\
\midrule
Baseline & 77.17 & 46.7 \\
SimData-Easy & 78.76 & 47.9 \\
SimData-Lite & 78.27 & 46.9 \\
SimData-Hard & 81.06 & 47.1 \\
SimData-Ultra & \textbf{89.63} & \textbf{48.7} \\
\bottomrule
\end{tabular}
\end{table}